\documentclass[10pt,twocolumn,letterpaper]{article}

\usepackage{wacv}              %
\makeatletter
\@namedef{ver@everyshi.sty}{
    \makeatother
    \usepackage{tikz}}

\usepackage{graphicx}
\usepackage{amsmath}
\usepackage{amssymb}
\usepackage{booktabs}
\usepackage{tabularx}
\usepackage[accsupp]{axessibility}

\usepackage[pagebackref,breaklinks,colorlinks]{hyperref}

\usepackage[capitalize]{cleveref}
\crefname{section}{Sec.}{Secs.}
\Crefname{section}{Section}{Sections}
\Crefname{table}{Table}{Tables}
\crefname{table}{Tab.}{Tabs.}

\usepackage{graphicx}
\usepackage{comment}
\usepackage{graphbox}  %
\usepackage{amsmath}
\usepackage{amssymb}
\usepackage[inline]{enumitem}
\usepackage{booktabs}
\usepackage{microtype}
\usepackage{xspace}
\usepackage[dvipsnames]{xcolor}
\newcommand{\mypar}[1]{\vspace{2mm}\noindent\textbf{#1}}
\usepackage{tikz}
\usepackage{numprint}

\usetikzlibrary{arrows}
\usetikzlibrary{backgrounds}
\usetikzlibrary{calc} 
\usetikzlibrary{positioning}
\usetikzlibrary{shapes.geometric}
\usetikzlibrary{fit}

\usepackage[pagebackref,breaklinks,colorlinks]{hyperref}

\usepackage[capitalize]{cleveref}
\crefname{section}{Sec.}{Secs.}
\Crefname{section}{Section}{Sections}
\Crefname{table}{Table}{Tables}
\crefname{table}{Tab.}{Tabs.}

\definecolor{indiagreen}{HTML}{138808}%
\definecolor{papaya}{HTML}{EE892F}%

\newcommand{\enc}{\mathsf{enc}}
\newcommand{\dec}{\mathsf{dec}}

\newcommand{\mb}{\mathbf{m}}
\newcommand{\xb}{\mathbf{x}}

\newcommand{\deemph}[1]{\textcolor{gray}{#1}}
\newcommand{\cfg}[1]{\textsf{\textsc{#1}}}

\newcommand{\grad}{\texttt{GradCAM}\xspace}
\newcommand{\att}{\texttt{Attention}\xspace}
\newcommand{\patch}{\texttt{Patches}\xspace}

\def\wacvPaperID{841} %
\def\confName{WACV}
\def\confYear{2024}

\begin{document}

\title{Weakly-supervised deepfake localization in diffusion-generated images}

\author{Dragoș-Constantin Țânțaru\\
Bitdefender\\
{\tt\small dtantaru@bitdefender.com}
\and
Elisabeta Oneață\\
Bitdefender\\
{\tt\small eoneata@bitdefender.com}
\and
Dan Oneață \\
University Politehnica of Bucharest\\
{\tt\small dan.oneata@gmail.com}}

\maketitle

\begin{abstract}
The remarkable generative capabilities of denoising diffusion models have raised new concerns regarding the authenticity of the images we see every day on the Internet.
However, the vast majority of existing deepfake detection models are tested against previous generative approaches (e.g. GAN) and usually provide only a  ``fake'' or ``real'' label per image.
We believe a more informative output would be to augment the per-image label with a localization map indicating which regions of the input have been manipulated.
To this end, we frame this task as a weakly-supervised localization problem 
and identify three main categories of methods (based on either explanations, local scores or attention), which we compare on an equal footing by using the Xception network as the common backbone architecture.
We provide a careful analysis of all the main factors that parameterize the design space:
choice of method, type of supervision, dataset and generator used in the creation of manipulated images;
our study is enabled by constructing datasets in which only one of the components is varied. 
Our results show that weakly-supervised localization is attainable, with the best performing detection method (based on local scores) being less sensitive to the looser supervision than to the mismatch in terms of dataset or generator.
\end{abstract}

\section{Introduction}
\label{sec:intro}

Image generation is improving by the day and it is arguably past the point where it is possible to perceptually distinguish between generated (fake) and real content.
Generative adversarial models (GAN) \cite{goodfellow2020generative},
normalizing flows \cite{rezende2015variational},
denoising diffusion probabilistic models (DDPM) \cite{pmlr-v37-sohl-dickstein15}---%
all provide excellent means for the creation of digital art or entertainment content.
However, the advances in image generation come at the cost of also easing malicious use, \eg, by altering reality or spreading misinformation.
To counter these harmful effects, deepfake detection methods 
are developed to discriminate between fake and real samples \cite{mirsky2021creation,nguyen2022cviu,verdoliva2020media}.

Among the classes of generative models, diffusion models are emerging as the dominant paradigm \cite{dhariwal2021diffusion},
showcasing impressive results on a wide array of tasks including
text-controlled image generation \cite{ramesh2022hierarchical,saharia2022photorealistic,rombach2022high,zhang2023controlnet} or
image-to-image translation \cite{saharia2022palette,rombach2022high,Lugmayr_2022_CVPR,zhang2023controlnet}.
Prior work on deepfake detection has naturally mostly considered detecting content generated by GANs 
\cite{Wang_2020_CVPR,gragnaniello2021gan,marra2019gans,Yu_2019_ICCV,chai2020makes},
but the computer vision community is now starting to consider DDPMs \cite{ricker2022towards,corvi2022detection}.
Here we continue this direction, going one step further to address the task of weakly-supervised deepfake localization.

First, we extend prior approaches to localise the manipulated area
and not only label the entire image as fake or real.
The binary output of the typical deepfake detection methods provides only coarse and opaque information,
especially in the frequent case of local manipulations and forgeries.
In this scenario, we would be much better served by a richer representation that could pinpoint which part of the image is likely to have been generated.
Another benefit of localization is that it allows the end-user to take more informed decisions.
For example,
changing the color of one's eyes may just be an innocuous enhancement of the user's appearance,
but the alteration of the movement of the lips in a video may hint towards a malicious use.
Instead of deciding upfront what is deemed to be fake or real, a localization method can defer this decision to the end user,
who is more informed and can tailor the method to their use case.

Second, in contrast to prior work, which addresses localization in a fully-supervised setting \cite{li2021noise,zhou2018learning,wu2019mantra,hu2020span}, we consider a weakly-supervised scenario, where we assume that we only have access to image-level labels
and the models are not explicitly trained for localization.
This setup is motivated by the fact that
generative methods are usually first developed in the context of full-image synthesis, and only then extended to the more specific cases of local editing, such as inpainting or attribute manipulation.
Moreover, ground truth manipulation masks might not always be available, especially for newly developed local manipulation methods.
Training a deepfake localization method in a weakly-supervised fashion (based on a global label) would allow us to be one step ahead of the potentially harmful uses involving local changes.

Our work brings the following contributions:
\begin{enumerate}
\item We propose a \textbf{weakly-supervised} framework for deepfake localization in images that allows to systematically uncover the importance of various factors (\textbf{model}, \textbf{supervision type}, \textbf{dataset}, \textbf{generator}) in the context of weakly-supervised localization of face manipulations.  
\item We generate a detailed dataset (more than 125k images) with locally- and fully-manipulated images that allows the \textbf{disentanglement} of different factors in deepfake manipulation localization.
The images are obtained using either newly introduced state-of-the-art generative models or a novel inpainting approach that incorporates a pretrained LDM \cite{rombach2022high} model in a diffusion-based inpainting method \cite{Lugmayr_2022_CVPR}. 
\item  We provide \textbf{extensive quantitative and qualitative results}
to understand the fundamental factors underlying the performance of weakly-supervised localization models.
Our analysis reveals the severity of out-of-domain degradation,
provides insights into the model's sensitivity to looser supervision or dataset mismatch,
and quantifies the performance across multiple classes of generative models.
Our code and dataset are available at {\small\url{https://github.com/bit-ml/dolos}}.
\end{enumerate}

\section{Related Work}

\mypar{Deepfake detection of GAN content.} There is a vast and continuously-growing body of work dedicated to the detection of GAN-generated images, see \cite{verdoliva2020media, mirsky2021creation, malik2022deepfake, nguyen2022cviu} for reviews.
Prior research has revealed many particularities of GAN content \cite{gragnaniello2021gan, Yu_2019_ICCV,cozzolino2021towards},
an important observation being the appearance of a fingerprint---an imperceptible pattern, which allows the identification of the GAN method and training dataset \cite{marra2019gans,Yu_2019_ICCV}.
Wang \etal \cite{Wang_2020_CVPR} also observe that all CNN-generated images share common systematic artifacts,
that can be easily picked up by a classifier, while in \cite{gragnaniello2021gan} the authors indicate that downsampling might destroy these high-frequency artifacts, which are the key to detection.

\mypar{Deepfake detection of DDPM content.}
Preliminary works on detecting diffusion-generated images made use of high-level cues such as inconsistencies in lighting \cite{farid2022lighting} or perspective distortion \cite{farid2022perspective}.
However, more common end-to-end detection networks were also tested on diffusion images  \cite{ricker2022towards,corvi2022detection},
focusing on the transferability across classes of generative models (from GAN to DDPM, and vice versa).
The prevailing observation is that detectors trained on one type of data do not generalize well to the other, but finetuning helps.

\mypar{Local manipulations.}
A common setup in deepfake creation is altering a person's face by reenactment, replacement, editing or synthesis using techniques known as face swap, face transfer, facial attribute manipulations or inpainting \cite{mirsky2021creation}.
These approaches result in local manipulations and are traditionally GAN-based.
Increasingly larger and more complex datasets and challenges have emerged
\cite{rossler2019faceforensics, jiang2020deeperforensics,li2020celeb,dolhansky2020deepfake,khalid2021fakeavceleb,he2021forgerynet}
and, with these, a considerable effort has been made to expose those types of fakes %
\cite{dong2022protecting,malik2022deepfake,haliassos2021lips,afchar2018mesonet,bappy2019hybrid,zhou2017two}.
However, actually localizing manipulations has arguably received less attention than detecting whether an image is fake or not.
Works that tackle localization rely on
local noise fingerprint patterns \cite{zhou2018learning, li2021noise,guillaro2022trufor,mareen2022comprint},
attention mechanisms \cite{dang2020detection,das2022gca,mazaheri2019cvprw} or
self-consistency checks \cite{huh2018fighting,agrawal2022cvprw}.
Very recent, concurrent works proposed a forensic framework for general manipulation  localization \cite{guillaro2022trufor} and a hierarchical fine-grained formulation for image forgery detection \cite{guo2023hierarchical}. Similar to us they consider diffusion-generated data with local forgeries, but differently they assume full supervision.

\section{Methodology}
\label{sec:methodology}

We first describe the methods used for deepfake detection and weakly-supervised localization (\S\ref{subsec:detection-methods}).
Then we detail the generative techniques that we are interested in detecting (\S\ref{subsec:generation-method})
and the datasets generated with these methods (\S\ref{subsec:datasets}).

\subsection{Methods for detection and localization}
\label{subsec:detection-methods}

\begin{figure}
    \centering
    \input{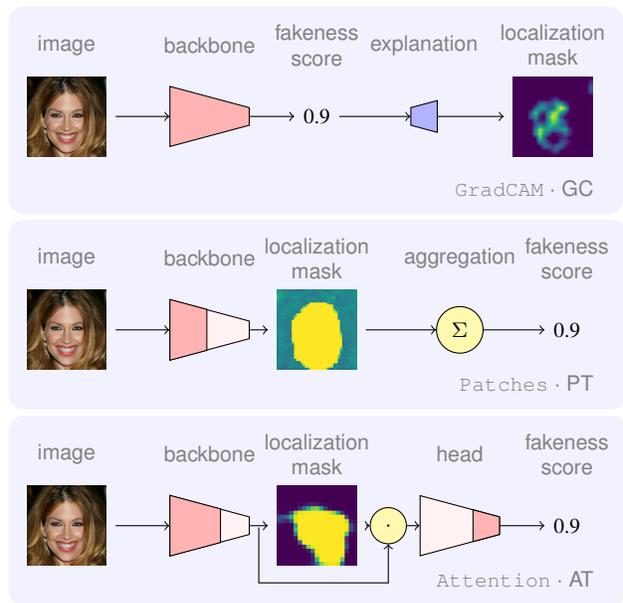}
    \caption{%
        Overview of the three types of approaches proposed for the detection and localization of deepfakes.
        Each method is able to produce a fakeness score (for detection) and a mask (for localization);
        the mask is obtained either explicitly (for the first model) or implicitly (for the second and third models).
    }
    \label{fig:methods-overview}
\end{figure}

The task of deepfake detection consists of predicting whether an image is either real or fake.
This task is usually framed as a binary classification problem and it is addressed using standard classification networks.
In this paper we are interested in evaluating the capabilities of such methods in a weakly-supervised setting:
if we assume only image-level labels,
can these classifiers be successfully used for \textit{localization} of partially manipulated images?

We identify and investigate three categories of architectures suitable for weakly-supervised localization.
These methods are based on either
explanations (\grad),
local scores (\patch)
or
attention (\att)
(for visual depictions see Figure \ref{fig:methods-overview}).
The first category is a general technique that given a trained classification network it uses explainability techniques to highlight the most predictive regions for the ``fake'' label.
The other categories implicitly construct the localization maps:
\patch produces local patch scores that are then used for classification,
while \att predicts an activation map that is used to pool relevant classification features.
To allow for a fair comparison we fix the backbone and, in particular,  we select the Xception network \cite{chollet2017xception}, which
has been shown to yield excellent results for deepfake detection of faces \cite{rossler2019faceforensics}.

The proposed methods
are inspired by and build upon state-of-the art deepfake detection methods \cite{rossler2019faceforensics,chai2020makes,dang2020detection},
but we %
further modify them as described below.

\mypar{\grad.} While GradCAM explanations were previously used in the deepfake detection literature \cite{zhou2017two,shiohara2022detecting,xu2023wacv,Boyd_2023_WACV}, they were mostly shown as qualitative results and rarely (if ever) evaluated quantitatively, in terms of how well they localize the input alterations.
In this paper we aim to quantify their performance and contrast them with other weakly-supervised localization methods.
Concretely, we endow the Xception~\cite{chollet2017xception} network with localization capabilities by applying GradCAM \cite{selvaraju2017grad} on the activations produced by block 11, the one before the last downsampling operation. 

\mypar{\patch.} We use Patch--Forensics~\cite{chai2020makes}, which 
is a truncated image classification network:
it takes the feature activations after a few layers and projects them to a patch-level score using 1 $\times$ 1 convolutions.
At train time, the loss is computed for each patch,
while at test time, it produces a detection score by averaging the per-patch softmax scores.
The authors experiment with two backbones (Xception \cite{chollet2017xception} and ResNet \cite{he2016deep}) and vary the number of %
layers that are kept.
We chose the Xception backbone truncated after the second block of layers, as this combination was shown to yield good performance \cite{chai2020makes}.
One advantage of \patch is that its output naturally corresponds to a localization map. %
While visualizations of the activation maps were shown in the original work, the localization performance was not quantified.

\mypar{\att.} We start from \cite{dang2020detection} which
augments an Xception \cite{chollet2017xception} backbone with a learned attention mask that is used to modulate the feature maps produced by the network.
The network is trained in a multi-task setting,
with a loss on the full-image fakeness score and
another one on the localization mask.
In the weakly-supervised scenario, when no groundtruth mask is provided,
the second term ensures that the maximum value of the predicted mask agrees with the image-level label.
We modify the original implementation in \cite{dang2020detection} to improve the performance and stabilize the training.
First, we replace the L1 loss on the mask with the binary cross-entropy loss (CE).
Second, we cross-validate the weight $\lambda$ that balances the two losses.
Our final loss is:
\begin{equation}
L = \mathrm{CE}(y, \hat{y}) + \lambda \mathrm{CE}(y, \max \hat{\mathbf{m}}),
\end{equation}
where $y$ is the true image label, $\hat y$ is the fakeness score and $\hat{\mathbf{m}}$ is the estimated localization mask.

\mypar{Fully-supervised localization.}
Along with the weakly-supervised setup we also consider the fully-supervised case to show an upper bound on the performance.
Since not all considered detection methods are able to be trained in a fully-supervised setting out of the box, we modify them to accommodate this setup:
for \grad  we truncate after block 11 and add a fully convolutional layer as in \cite{shelhamer2016fcn};
for \att we keep only the loss on the mask, otherwise the architecture remains the same;
for \patch we maintain the same architecture,
but instead of using the image label to supervise each feature prediction,
we use the downsized mask as groundtruth.

\subsection{Dataset generation methods }
\label{subsec:generation-method}

We use diffusion models to generate both full images and locally-inpainted ones.

\mypar{Diffusion denoising probabilistic models} (DDPM) \cite{pmlr-v37-sohl-dickstein15} are a class of generative models trained to reverse a diffusion process.
The forward diffusion process iteratively adds Gaussian noise to a sample until its distribution reaches a standard normal.
The reverse denoising process gradually removes noise, producing novel samples when starting from a random image.
The reverse process is implemented as a neural network (with parameters $\theta$) that predicts the mean $\mu_{\theta}(\xb_t, t)$ and covariance $\Sigma_{\theta}(\xb_t,t)$ of a Gaussian distribution:
\begin{align}
 p_{\theta}(\xb_{t-1}|\xb_t) = \mathcal{N}\left(\xb_{t-1};\mu_{\theta}(\xb_t, t), \Sigma_{\theta}(\xb_t,t)\right),
 \label{eq:predict}
\end{align}
where $\xb_t$ are images that are sequentially generated, from $t=T$ to $t=1$.

\mypar{Repaint: Inpainting with diffusion.}
The task of inpainting is to fill in the missing regions of an image $\xb_0$ such that the resulting composition looks natural;
the missing regions are usually specified by a binary mask $\mb$.
For inpainting with diffusion we use the approach of Lugmayr \etal \cite{Lugmayr_2022_CVPR}, whose method performs mask-guided decoding on any pretrained DDPM.
More precisely, at generation time they first sample a new image $\hat\xb_t$ from the previously-generated image, $\hat\xb_{t+1}$, according to Equation \eqref{eq:predict},
but then they replace the values of $\hat\xb_t$ outside the given mask $\mb$ with the values of the original image encoded after $t$ steps $\xb_t$: %
\begin{equation}
    \hat\xb_t \leftarrow \mb \odot \hat\xb_t + (1 - \mb) \xb_t
    \label{eq:repaint-schedule}
\end{equation}
This procedure ensures that the values outside the mask are preserved from the original image $\xb_0$.

\mypar{Repaint--LDM: Inpainting with diffusion in the latent space.}
Latent diffusion models (LDM) \cite{rombach2022high} have been shown to offer a scalable approach to generating high-fidelity images.
Their main idea consists of performing diffusion in the (low-dimensional) latent space of a variational autoencoder (VAE).
We translate this idea to inpainting by running the Repaint scheduler (Equation \ref{eq:repaint-schedule}) in the latent space, $\xb \leftarrow \enc(\xb)$, of the variational autoencoder and using an appropriately downsized mask, $\mb \leftarrow \mathsf{resize}(\mb)$.
This procedure generates an (inpainted) latent code, $\hat\xb$, which is then inverted to the original pixel space using the decoder of the VAE, $\dec(\hat\xb)$.
Notably, this method allows us to inpaint an image using any existing pretrained LDM model.
To the best of our knowledge, this approach to inpainting is novel.

\subsection{Datasets}
\label{subsec:datasets}

\begin{figure}%
    \def\hf{34pt}
    \scriptsize 
    \begin{center}        
    \setlength{\tabcolsep}{2pt}
        \begin{tabular}{c ccc ccc}
        & \multicolumn{6}{c}{Full image synthesis $\cdot$ generator: P2 \cite{choi2022perception}} \\
        \rotatebox[origin=c]{90}{\tiny CelebA-HQ}
        & \includegraphics[align=c,height=\hf]{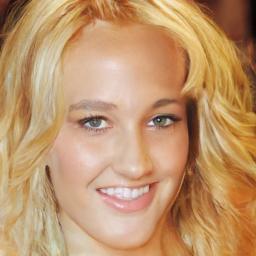}
        & \includegraphics[align=c,height=\hf]{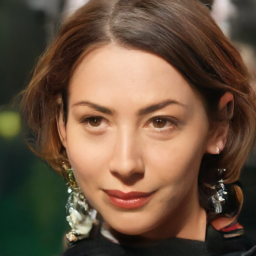}
        & \includegraphics[align=c,height=\hf]{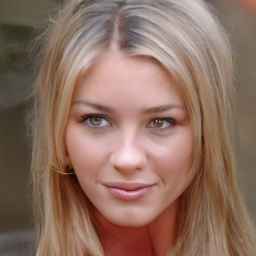}
        & \includegraphics[align=c,height=\hf]{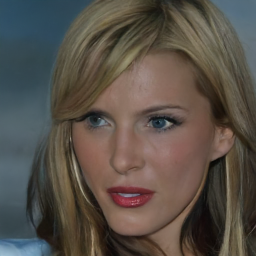}
        & \includegraphics[align=c,height=\hf]{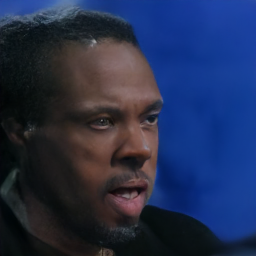}
        & \includegraphics[align=c,height=\hf]{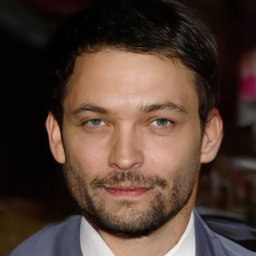}
        \vspace{3pt}
        \\
        \rotatebox[origin=c]{90}{\tiny FFHQ}
        & \includegraphics[align=c,height=\hf]{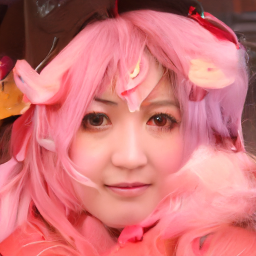}
        & \includegraphics[align=c,height=\hf]{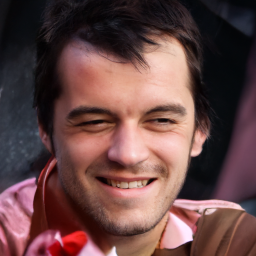}
        & \includegraphics[align=c,height=\hf]{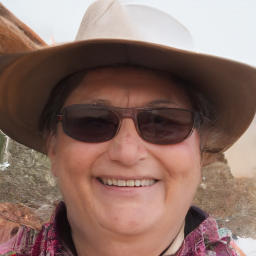}
        & \includegraphics[align=c,height=\hf]{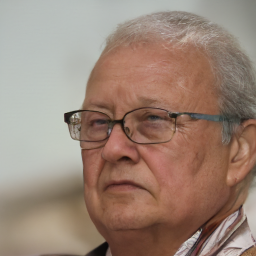}
        & \includegraphics[align=c,height=\hf]{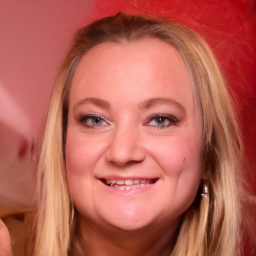}
        & \includegraphics[align=c,height=\hf]{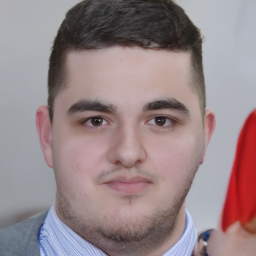}
        \vspace{1pt}
        \\
        \midrule
        & \multicolumn{6}{c}{Local manipulations $\cdot$ dataset: CelebA-HQ} \\
        & \includegraphics[align=c,height=\hf]{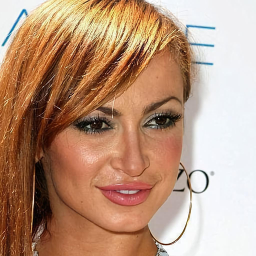}
        & \includegraphics[align=c,height=\hf]{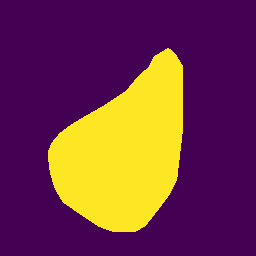}
        & \includegraphics[align=c,height=\hf]{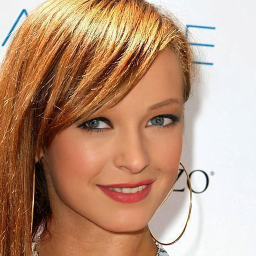}
        & \includegraphics[align=c,height=\hf]{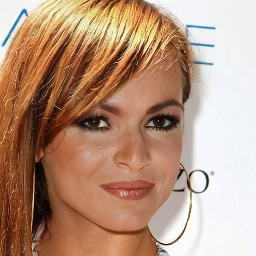}
        & \includegraphics[align=c,height=\hf]{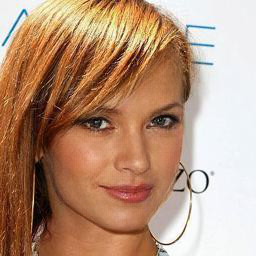}
        & \includegraphics[align=c,height=\hf]{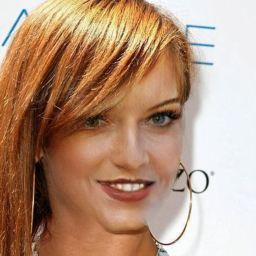}
        \vspace{3pt}
        \\
        & \includegraphics[align=c,height=\hf]{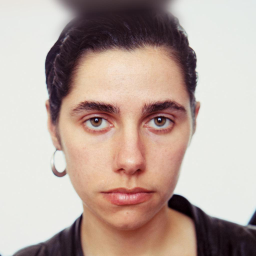}
        & \includegraphics[align=c,height=\hf]{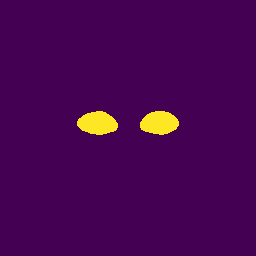}
        & \includegraphics[align=c,height=\hf]{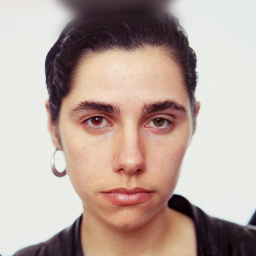}
        & \includegraphics[align=c,height=\hf]{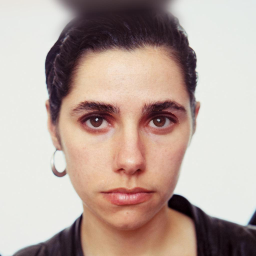}
        & \includegraphics[align=c,height=\hf]{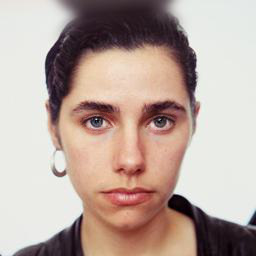}
        & \includegraphics[align=c,height=\hf]{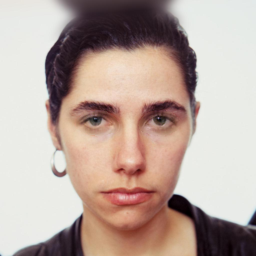}
        \\
        & original
        & mask
        & Repaint
        & Repaint
        & LaMa
        & Pluralistic
        \\
        & 
        & 
        & P2
        & LDM
        & 
        &
    \end{tabular}            
    \end{center}
    \caption{%
        Examples from our generated dataset.
        The first two rows represent fully-generated images using P2 on CelebAHQ and FFHQ, respectively. 
        The last two rows represent locally-inpainted images using Repaint--P2, Repaint--LDM, LaMa, Pluralistic, respectively;
        notice the high realism of images obtained with both large and small masks.
     }
    \label{fig:example_dataset}
\end{figure}

\begin{table}
\tabcolsep 3pt
\centering
\footnotesize
\begin{tabular}{l l l l rrr}
\toprule
& \multicolumn{2}{c}{Generator} & & \multicolumn{3}{c}{Num. samples} \\
\cmidrule(lr){2-3}
\cmidrule(lr){5-7}
Type       & Family       & Model        &  Dataset & Train & Val. &  Test \\
\midrule                                                                  
Real       & --           &           -- & CelebA &  9k &  900  & 900  \\
Real       & --           &           -- &   FFHQ &  9k &  900  & --   \\
\midrule
Fake full  & Diffusion    & P2           & CelebA &  9k &   1k  & --   \\
Fake full  & Diffusion    & P2           &   FFHQ &  9k &   1k  & --   \\
\midrule
Fake local & Diffusion    & Repaint--P2  & CelebA & 30k &   3k  & 8.5k \\
Fake local & Diffusion    & Repaint--P2  &   FFHQ & 30k &   3k  & --   \\
Fake local & Latent diff  & Repaint--LDM & CelebA &  9k &  900  & 900  \\
Fake local & Fourier conv & LaMa         & CelebA &  9k &  900  & 900  \\
Fake local & GAN          & Pluralistic  & CelebA &  9k &  900  & 900  \\
\bottomrule
\end{tabular}
\caption{%
     Details of our proposed dataset, which contains locally- and fully-generated images from multiple types of generators.
     The dataset is designed to allow for a principled analysis of %
     multiple factors: manipulation type, generator, source dataset.
     We provide:
     (i) fully-generated images on CelebA-HQ and FFHQ using P2 \cite{choi2022perception};
     (ii) locally-inpainted images on FFHQ using Repaint--P2 and 
     on CelebA-HQ using Repaint--P2, Repaint--LDM, Pluralistic\cite{zheng2019pluralistic}, LaMa\cite{suvorov2022resolution} (using the same masks).
}
\label{tab:gen_datasets}
\end{table}

To train and evaluate our models,
we use real images and two types of fake images:
fully-synthesized and locally-manipulated images.
The datasets are summarized in Table~\ref{tab:gen_datasets} and
examples are shown in Figure~\ref{fig:example_dataset}.

\mypar{Real data.}
We use the CelebA-HQ and FFHQ face datasets as sources of real data.
CelebA-HQ \cite{karras2017progressive} consists of 30k images that were selected and processed from the CelebA dataset \cite{liu2015deep};
we keep the original splits for training, validation and testing.
FFHQ \cite{karras2019style} consists of 70k PNG images that have been crawled from Flickr and automatically aligned and cropped.
Both datasets are popular choices for training generative models and, consequently, are suitable choices for training deepfake detection models.
We select a subset of 9k train and 900 validation images from each of the two datasets to match the number of fake images that are generated.

\mypar{Fake data: Full-image synthesis.}
We use the perception-prioritized (P2) diffusion method of Choi \etal \cite{choi2022perception} to sample fully-synthetic images.
We chose this approach because
(i) the authors provide pretrained models on the two real datasets mentioned above (CelebA-HQ and FFHQ), which enable a systematic experimentation, and
(ii) the models are lightweight and hence the inference is reasonably fast.
For both datasets we sample 10k images: 9k for training and 1k for validation.
We do not evaluate on these fully-synthesized sets, hence no test set is provided. 
We refer to these datasets as P2/CelebA-HQ and P2/FFHQ, respectively.

\mypar{Fake data: Local manipulations.}
We generate two locally-manipulated datasets using the Repaint method \cite{Lugmayr_2022_CVPR} to inpaint images from the CelebA-HQ and FFHQ datasets.
We use the Repaint method on top of pretrained P2 models, namely its variants trained on CelebA-HQ and FFHQ, respectively.
The inpainted regions correspond to various face attributes (skin, hair, eyes, mouth, nose, glasses).
For CelebA-HQ, these annotations were manually labeled and are available in the CelebAMask-HQ \cite{karras2017progressive} extension of the dataset, while for FFHQ these are obtained using a pretrained face segmentation method \cite{CelebAMask-HQ}.
Given an image (corresponding to the identity of a person) we generate multiple inpaintings by randomly sampling masks corresponding to these face attributes and, for the smaller parts (eyes, mouth, nose), by also dilating them with a kernel of randomly-chosed size, but up to 15 pixels.
We refer to the resulting datasets as
Repaint--P2/CelebA-HQ and Repaint--P2/FFHQ;
the former will represent our main test bed,
while the latter is used only at training. %

To be able to systematically study the importance of the generator we inpainted a subset of 9k images used in Repaint--P2/CelebA-HQ with three other methods:
Repaint-LDM (ours), LaMa \cite{suvorov2022resolution}, Pluralistic \cite{zheng2019pluralistic}.
Repaint-LDM adapts the Repaint method to operate in the latent space by using the LDM model \cite{rombach2022high};
LaMa is an inpainting method that uses an autonecoder with Fourier convolutions \cite{chi2020fourierconv};
Pluralistic is a conditional variational autoencoder with adversarial loss.
We have chosen these methods since they all provide pretrained models on the CelebA-HQ dataset.
This allows us to inpaint the same images using the same masks, and isolate the differences attributed to the change of generator.

\section{Experimental setup}

\mypar{Implementation details.} Following the recommendation of Chai \etal \cite{chai2020makes}, we ensure that real and fake images both follow exactly the same preprocessing steps prior to passing them through the detection methods.
These steps include the input resolution and resize algorithm.
Consequently, we process both CelebA-HQ and FFHQ images as they were processed for training the generator,
that is, we resize them to 256 $\times$ 256 using bicubic interpolation.

\mypar{Tasks and metrics.}
Localization is the main task that we tackle.
We report intersection over union (IoU) and pixel-wise binary classification accuracy (PBCA).
These metrics assume binary prediction and we use a fixed threshold of 0.5 for binarization.
The detection methods generate masks of different sizes:
19 $\times$ 19 for \grad and \att,
37 $\times$ 37 for \patch.
For a fair evaluation we resize them to the size of the input image: 256 $\times$ 256.

We also report results on detection, the task of telling apart fake images from real images.
We rank the images by their per-image fakeness score, which is output by each method as illustrated in Figure~\ref{fig:methods-overview}.
The detection performance is then measured in terms of average precision (AP), which is a threshold-free metric.

\section {Experiments}
\label{sec:experiments}

Our experiments evaluate the proposed methods with different levels of supervision, gradually changing the dataset and the generators in order to quantify their importance for localization. 
We investigate the performance using three main levels of supervision:

\begin{itemize}
    \item \textbf{Setup \cfg{a} (label \& full)}
        is a weakly-supervised setup in which we have access to fully-generated images as fakes and, consequently, only image-level labels.
        We use 9k fake images, fully synthesized by P2, and 9k real images from the corresponding dataset on which P2 was trained.
    \item \textbf{Setup \cfg{b} (label \& partial)}
        is a weakly-supervised setup in which we have access to partially-manipulated images, but only with image-level labels (no localization information).
        This means that while an image may be labelled as ``fake'', not all of its regions are fake.
        We use 9k locally-modified images by Repaint--P2 and 9k real images from the corresponding training dataset.
    \item \textbf{Setup \cfg{c} (mask \& partial)}
       is a fully-supervised setting, in which we have access to ground-truth localization masks of partially-manipulated images.
       We uses 30k locally-modified images by Repaint--P2; for this setup, no real images are used.
\end{itemize}

To evaluate localization we use 8.5k locally-manipulated images produced by Repaint--P2/CelebA-HQ
and to evaluate detection we use 900 real images from CelebA-HQ and 900 fakes from Repaint--P2/CelebA-HQ.
Note that the evaluation is carried on the same data regardless of the setup.
Table 1 from the supplementary material summarizes the data used in each of the three setups.

\subsection{Evaluating localization abilities}
\label{subsec:quantifying-localization}

\begin{table*}
    \centering
    \small
    \begin{tabular}{c cc rrr rrr rrr}
        \toprule
        & & &
        \multicolumn{3}{c}{IoU (\%)} &
        \multicolumn{3}{c}{PBCA (\%)} &
        \multicolumn{3}{c}{AP (\%)} \\
        \cmidrule(lr){4-6}
        \cmidrule(lr){7-9}
        \cmidrule(lr){10-12}
        \color{gray} setup & sup. & generator & 
        \multicolumn{1}{c}{GC} & \multicolumn{1}{c}{PT} & \multicolumn{1}{c}{AT} & 
        \multicolumn{1}{c}{GC} & \multicolumn{1}{c}{PT} & \multicolumn{1}{c}{AT} & 
        \multicolumn{1}{c}{GC} & \multicolumn{1}{c}{PT} & \multicolumn{1}{c}{AT} \\
        \midrule                           
        \deemph{\cfg{a}} & label & full    & 16.8 & \bf 64.9 &  9.7 & 83.1 & \bf 96.7 &     83.4 & 67.3 & \bf 95.3 & 79.3 \\
        \deemph{\cfg{b}} & label & partial & 21.5 & \bf 37.7 & 23.2 & 85.1 &     79.8 & \bf 86.3 & 94.4 & \bf 95.3 & 94.4 \\
        \deemph{\cfg{c}} & mask  & partial & 83.7 & \bf 84.5 & 70.3 & 96.8 & \bf 98.6 &     97.6 &   -- &       -- &   -- \\
        \bottomrule
    \end{tabular}
    \caption{%
        Evaluation of the three selected localization techniques (\grad GC, \patch PT, \att AT)
        on the Repaint--P2/CelebA-HQ dataset
        using different levels of supervision:
        image-level label on full images (\cfg{a}),
        image-level label on locally manipulated images (\cfg{b}), and
        fully-supervised masks (\cfg{c}).
        We evaluate both localization (using IoU and PBCA) and detection (using AP).
        \patch systematically outperforms the other two methods under most of the scenarios and metrics.%
    }
    \label{tab:methods}
\end{table*}

\begin{figure}%
    \def\hf{34pt}
    \footnotesize
    \begin{center}        
    \setlength{\tabcolsep}{1.5pt}
        \begin{tabular}{c ccc cc ccc} %
            & real
            & inpainted
            & mask
            & &
            & real
            & inpainted
            & mask
            \\
            & \includegraphics[align=c,height=\hf]{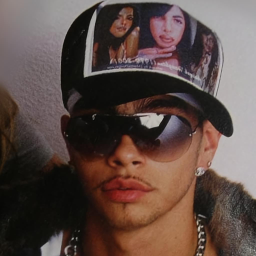}
            & \includegraphics[align=c,height=\hf]{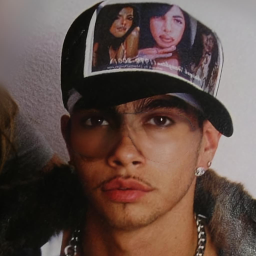}
            & \includegraphics[align=c,height=\hf]{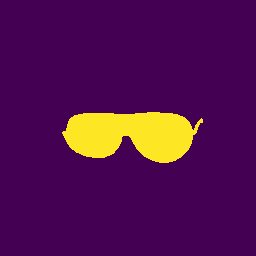}
            & &
            & \includegraphics[align=c,height=\hf]{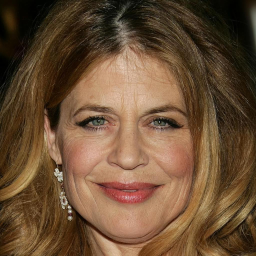}
            & \includegraphics[align=c,height=\hf]{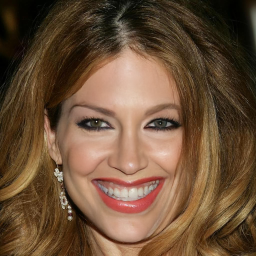}
            & \includegraphics[align=c,height=\hf]{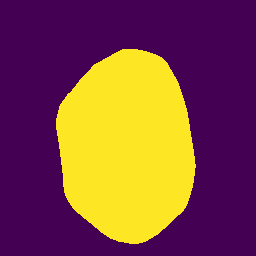}
            \vspace{3pt}
            \\
            & GC
            & PT
            & AT
            & &
            & GC
            & PT
            & AT
            \\
            \cfg{a} 
            & \includegraphics[align=c,height=\hf]{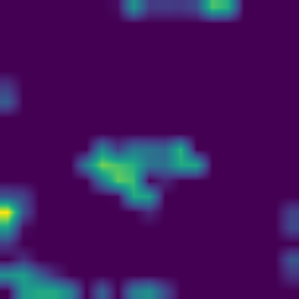}
            & \includegraphics[align=c,height=\hf]{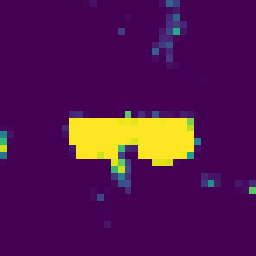}
            & \includegraphics[align=c,height=\hf]{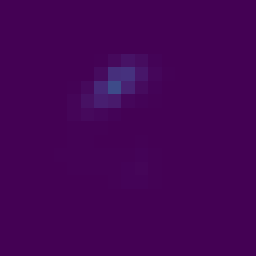}
            & &
            & \includegraphics[align=c,height=\hf]{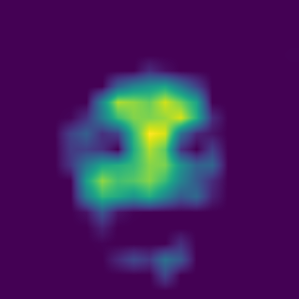}
            & \includegraphics[align=c,height=\hf]{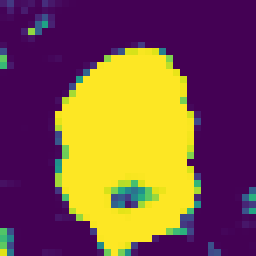}
            & \includegraphics[align=c,height=\hf]{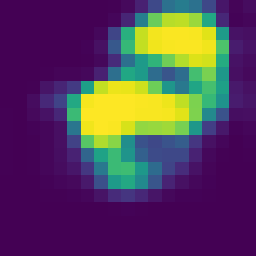}
            \vspace{3pt}
            \\
            \cfg{b} 
            & \includegraphics[align=c,height=\hf]{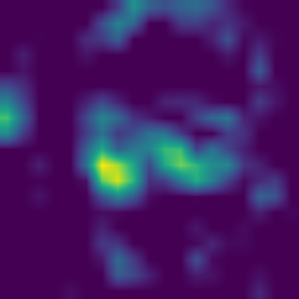}
            & \includegraphics[align=c,height=\hf]{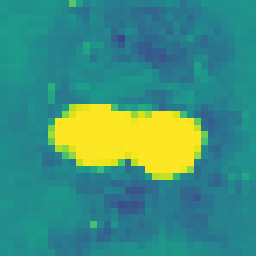}
            & \includegraphics[align=c,height=\hf]{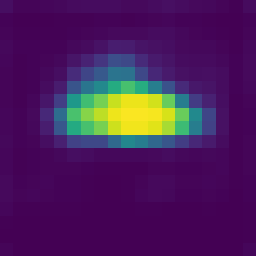}
            & &
            & \includegraphics[align=c,height=\hf]{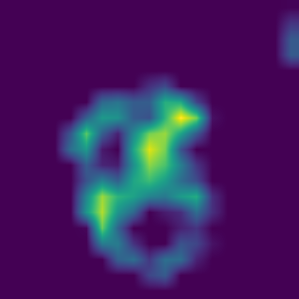}
            & \includegraphics[align=c,height=\hf]{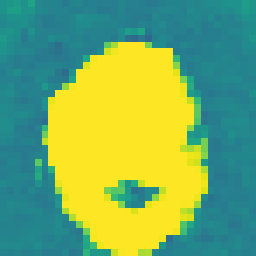}
            & \includegraphics[align=c,height=\hf]{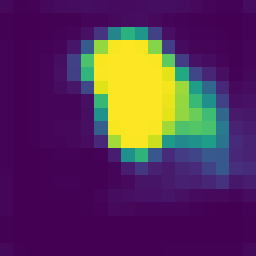}
            \vspace{3pt}
            \\  
            \cfg{c} 
            & \includegraphics[align=c,height=\hf]{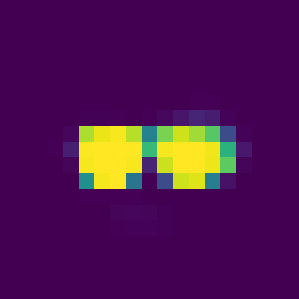}
            & \includegraphics[align=c,height=\hf]{imgs/maps/patches/train_celeba/setupA/27848-0.png}
            & \includegraphics[align=c,height=\hf]{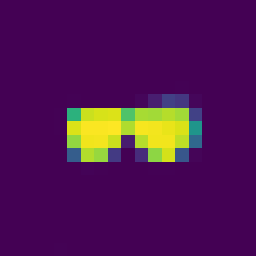}
            & &
            & \includegraphics[align=c,height=\hf]{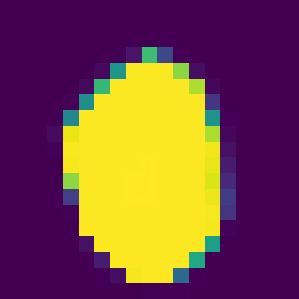}
            & \includegraphics[align=c,height=\hf]{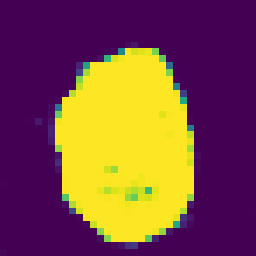}
            & \includegraphics[align=c,height=\hf]{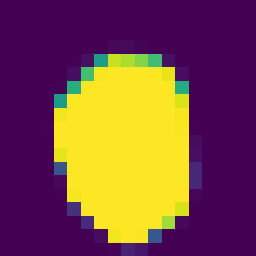}
        \end{tabular}            
    \end{center}
    \caption[localization visual results]{%
      Soft localization maps produced by the three approaches using different levels of supervision.
      \patch can accurately detect the manipulations after having seen only fully-generated fake images (scenario \cfg{a}) or locally-inpainted images with only image-level supervision (scenario \cfg{b}).
      Both \att and \grad struggle in scenarios \cfg{a} and \cfg{b}.
      All methods recover the manipulated region in the fully supervised scenario, \cfg{c}. This suggests that operating at a patch level is better suited for recovering local manipulations than either using GradCAM or attention.}
    \label{fig:maps}
\end{figure}

We evaluate all three proposed approaches for localization in the three setups described above.
To exclude other factors of variation we maintain the image generator and source dataset fixed, that is,
for scenario \cfg{a} we train on P2/CelebA-HQ, while
for scenarios \cfg{b} and \cfg{c} we use Repaint--P2/CelebA-HQ.
Real data from CelebA-HQ is used in setups \cfg{a} and \cfg{b},
while for the fully supervised scenario, setup \cfg{c}, real data is not needed.
Results for both localization and detection are shown in Table \ref{tab:methods}.

Among the selected methods, we see that \patch generally outperforms the other two approaches across multiple setups and metrics (bold values in Table~\ref{tab:methods}).
We see that localization performance is strong for all methods when training in the fully supervised scenario (setup \cfg{c}) and performance drops as we move to the two weakly supervised setups (setups \cfg{a} and \cfg{b}).
Interestingly, \grad and \att perform better in setup \cfg{b} than in setup \cfg{a},
while for \patch we observe the reverse trend.
We believe that \patch is worse in setup \cfg{b} %
because the loss is set at patch-level,
and the patch labels are inherently noisy as we use partially-manipulated images at input.

In terms of detection (the `AP' columns in Table~\ref{tab:methods}),
we observe strong performance of \patch in both weakly supervised setups, \cfg{a} and \cfg{b}.
Interestingly, the detection performance is good for all models in setup \cfg{b}.
In retrospect, this is expected since for the detection task in setup \cfg{b} the train data matches the test data.

Figure \ref{fig:maps} shows examples of the localization maps produced by the detection methods in all three scenarios.
We notice that \patch is able to partially recover the manipulated areas even in setups \cfg{a} and \cfg{b}. %
In setup \cfg{b} we observe that due to the noisy labels the model fires also on the background regions.
\grad and \att struggle more in the weakly-supervised scenarios and their outputs are qualitatively different:
the former seems to produce weaker activations, which are spread through irrelevant areas of the image (especially in scenario \cfg{a}),
while the latter produces less precise localizations.

\subsection{Generalization across source datasets}

\begin{table}
\tabcolsep 4pt
\centering
\small
\begin{tabular}{c cc rrr rrr}
\toprule
& & & \multicolumn{3}{c}{CelebA-HQ }  & \multicolumn{3}{c}{FFHQ } \\
\cmidrule(lr){4-6}
\cmidrule(lr){7-9}
& sup. & generator &
\multicolumn{1}{c}{IoU} & \multicolumn{1}{c}{PBCA} & \multicolumn{1}{c}{AP} &
\multicolumn{1}{c}{IoU} & \multicolumn{1}{c}{PBCA} & \multicolumn{1}{c}{AP} \\
\midrule                           
\deemph{\cfg{a}}  & label & full    & 64.9 & 96.7 & 95.3 & 25.1 & 88.9 & 84.4 \\
\deemph{\cfg{b}}  & label & partial & 37.7 & 79.8 & 95.3 & 23.3 & 64.4 & 75.2 \\
\deemph{\cfg{c}}  &  mask & partial & 84.5 & 98.6 &  --  & 32.3 & 89.2 & --   \\
\bottomrule
\end{tabular}
\caption{%
     Evaluation of \patch on the Repaint--P2/CelebA-HQ dataset using two training datasets: CelebA-HQ and FFHQ.
     When the source dataset does not match the target one,
     we observe a consistent drop in performance across all scenarios.
     This is more evident in scenario \cfg{b} where only image-level supervision is available for locally-manipulated images.  
}
\label{tab:datasets}
\end{table}

Generalization is a key desirable property of deepfake detectors.
Here, we assess how localization is affected by datasets shifts.
To this end, we design an experiment in which the training and testing data come from different source datasets, while fixing the generator and the detection method.
Training is either based on fake data derived from CelebA-HQ (P2/CelebA-HQ for scenario \cfg{a} and  Repaint--P2/CelebA-HQ for scenarios \cfg{b} and \cfg{c}) or FFHQ (P2/FFHQ for scenario \cfg{a} and Repaint--P2/FFHQ for scenarios \cfg{b} and \cfg{c}),
while the testing is carried on Repaint--P2/CelebA-HQ.

\begin{figure}%
    \def\hf{34pt}
    \footnotesize
    \begin{center}        
    \setlength{\tabcolsep}{1.5pt}
        \begin{tabular}{c cc c cc c cc}
            & inpainted
            & mask
            &
            & inpainted
            & mask 
            &
            & inpainted
            & mask 
            \\
            & \includegraphics[align=c,height=\hf]{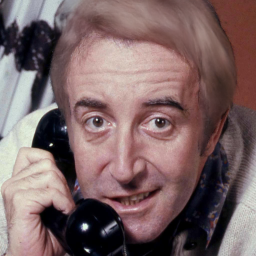}
            & \includegraphics[align=c,height=\hf]{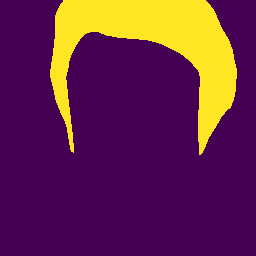}
            &
            & \includegraphics[align=c,height=\hf]{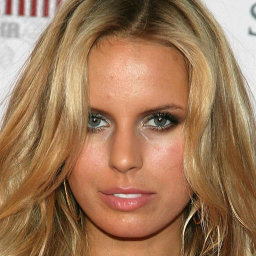}
            & \includegraphics[align=c,height=\hf]{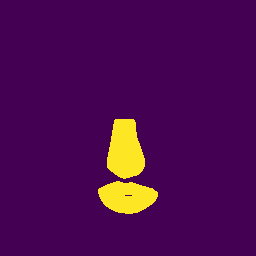}
            &
            & \includegraphics[align=c,height=\hf]{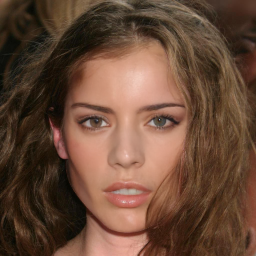}
            & \includegraphics[align=c,height=\hf]{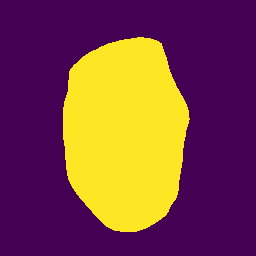}
            \vspace{3pt}
            \\
            &
            \multicolumn{2}{c}{train \textit{vs} test} & &
            \multicolumn{2}{c}{train \textit{vs} test} & &
            \multicolumn{2}{c}{train \textit{vs} test} \\
            & same
            & different
            &
            & same
            
            & different
            &
            & same
            & different
            \\
            \cfg{a} 
            & \includegraphics[align=c,height=\hf]{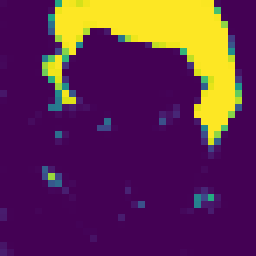}
            & \includegraphics[align=c,height=\hf]{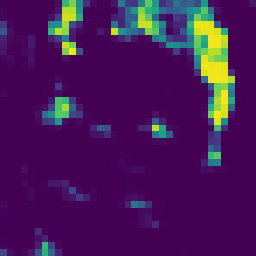}
            &
            & \includegraphics[align=c,height=\hf]{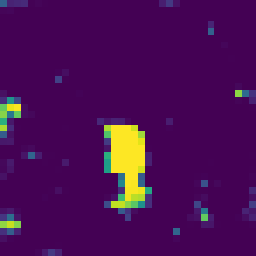}
            & \includegraphics[align=c,height=\hf]{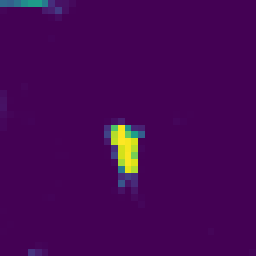}
            &
            & \includegraphics[align=c,height=\hf]{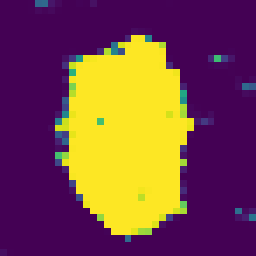}
            & \includegraphics[align=c,height=\hf]{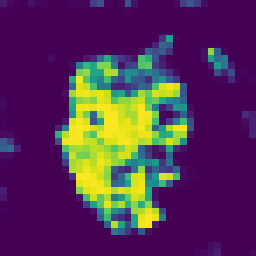}
            \vspace{3pt}
            \\
            \cfg{b} 
            & \includegraphics[align=c,height=\hf]{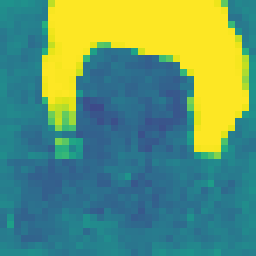}
            & \includegraphics[align=c,height=\hf]{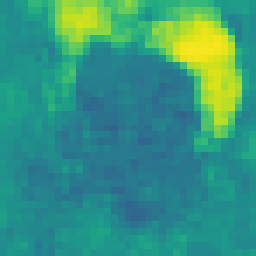}
            &
            & \includegraphics[align=c,height=\hf]{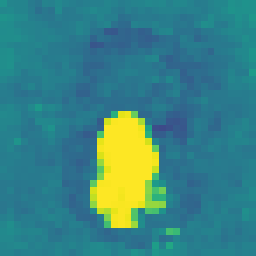}
            & \includegraphics[align=c,height=\hf]{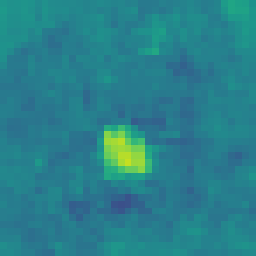}
            &
            & \includegraphics[align=c,height=\hf]{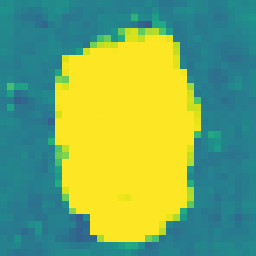}
            & \includegraphics[align=c,height=\hf]{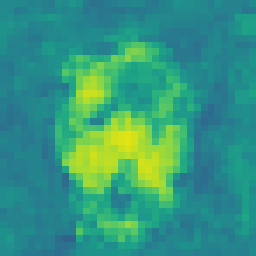}
            \vspace{3pt}
            \\  
            \cfg{c} 
            & \includegraphics[align=c,height=\hf]{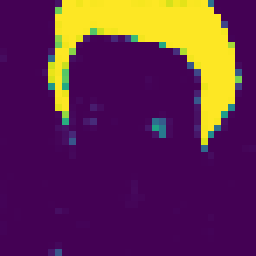}
            & \includegraphics[align=c,height=\hf]{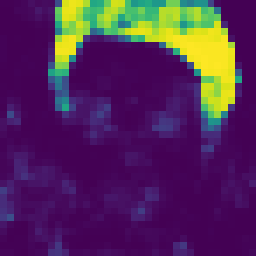}
            &
            & \includegraphics[align=c,height=\hf]{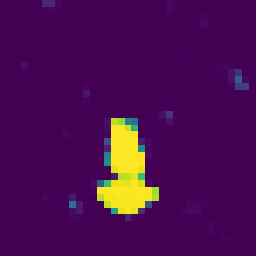}
            & \includegraphics[align=c,height=\hf]{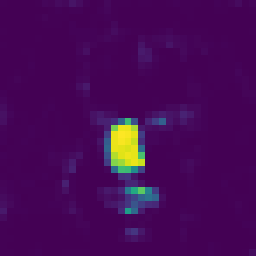}
            &
            & \includegraphics[align=c,height=\hf]{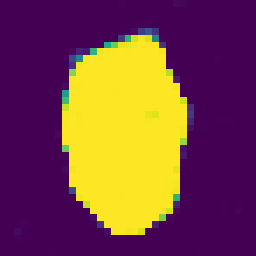}
            & \includegraphics[align=c,height=\hf]{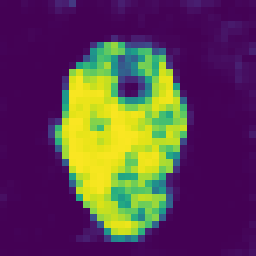}
            \vspace{3pt}
        \end{tabular}
    \end{center}
    \caption[localization visual results]{%
        Soft localization maps produced by \patch when using the same \textit{vs} different source datasets for training and testing.
        For training we use data derived form either CelebA-HQ or FFHQ,
        while for testing we use data derived from CelebA-HQ.
        When there is a dataset mismatch (the `different' column), we observe maps that are less sharp and eroded,
        especially in the weakly supervised scenarios, \cfg{a} and \cfg{b}.
        The noisy training of scenario \cfg{b} dims the separation between real and fake regions.
    }
    \label{fig:maps_diff_datasets}
\end{figure}

Quantitative results are shown in Table \ref{tab:datasets} for all scenarios under both localization and detection metrics.
We observe a consistent drop in performance across all scenarios and metrics when there is a dataset mismatch.
A closer look at the soft localization maps reveals a more complete picture.
The `different' columns in Figure \ref{fig:maps_diff_datasets}
show that training on FFHQ still produces qualitatively reasonable predictions even for small regions (nose and mouth).
However, in this mismatched setting the predictions are less certain at the boundaries and the masks appears to be eroded or with holes.

\begin{figure}
    \centering
    \includegraphics[width=0.48\textwidth]{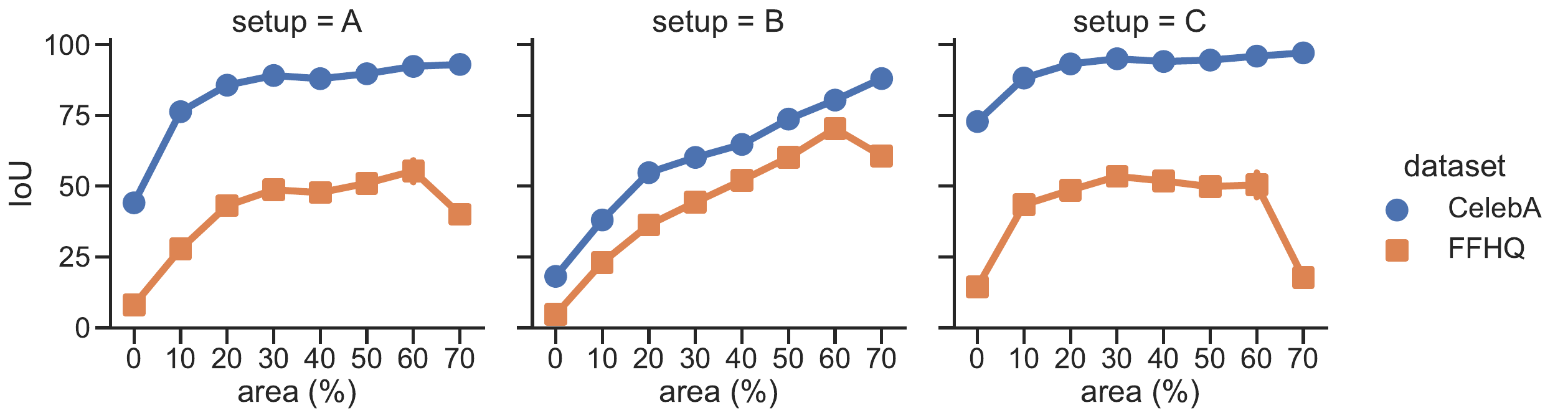}
    \caption{%
        IoU as a function of the manipulated area (as percentage) for all three setups when changing the training dataset:
        CelebA-HQ (same as test; blue) or FFHQ (different from test; orange).
    }
    \label{fig:score-vs-mask-area}
\end{figure}

To better assess the estimated localization maps we look at the how their accuracy varies with the size of the manipulated region.
In Figure \ref{fig:score-vs-mask-area} we show the IoU score as a function of the mask area for the three setups when
(i) the training dataset matches the one at test time (blue line), and
(ii) is different (orange line).
We observe that larger manipulations are generally easier to correctly locate (increasing IoU with area) and that the dataset mismatch results in a sizable drop in performance.
However, for setup \cfg{b} the slopes are much steeper and the gap between the two curves is reduced.
We believe that this happens because in this setup the model fires also on the background and, as the background takes most of the image, this region will impact most of the performance.

\subsection{Generalization across generators}
\label{susbec:cross-generator-performance}

\begin{figure}%
    \includegraphics[width=0.5\textwidth,trim={1.75cm 0.1cm 1.5cm 0cm},clip]{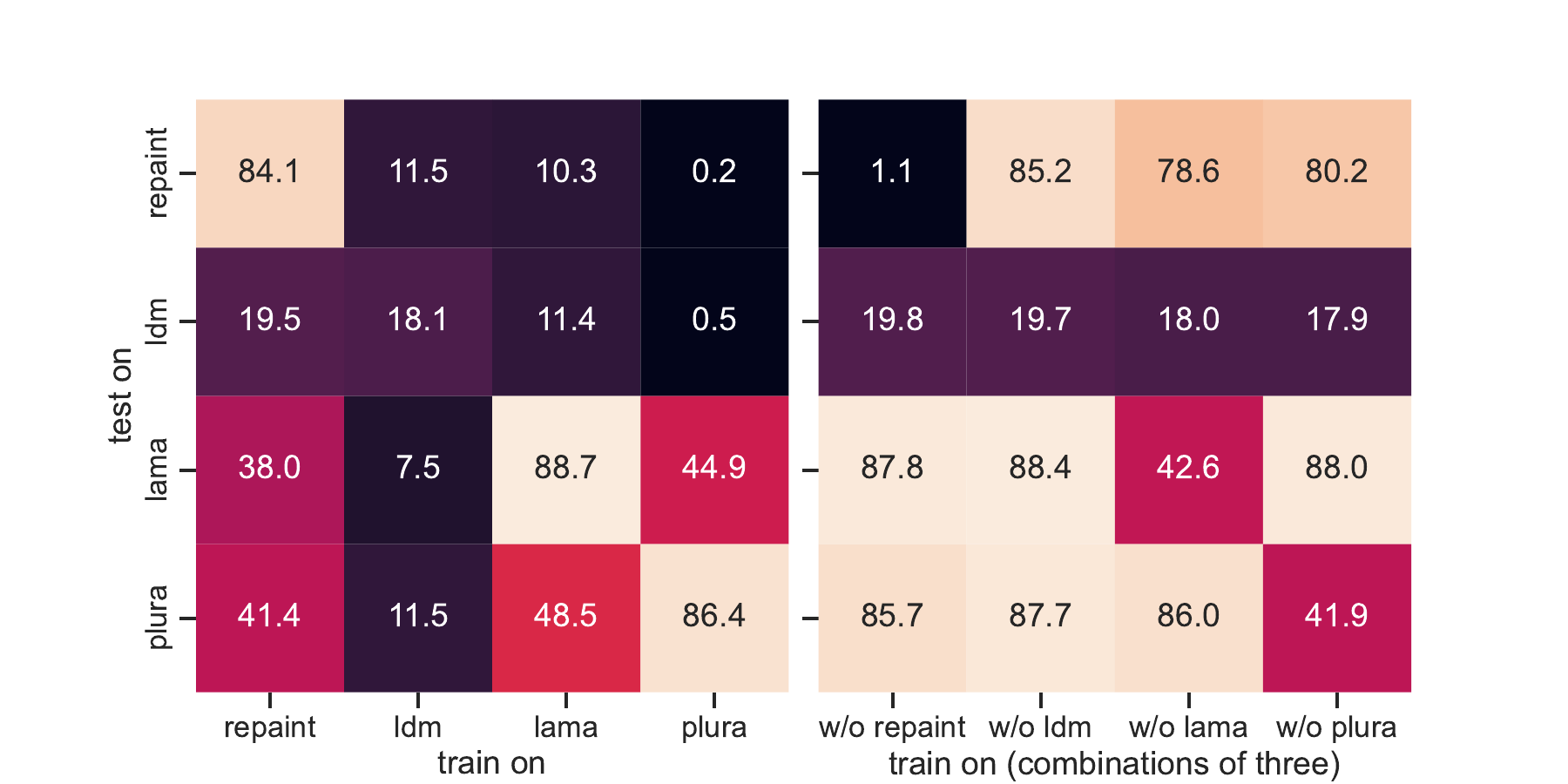}
    \caption{%
        Localization performance (IoU) across four inpainting methods (Repaint, LDM, LaMa, Pluralistic) and their combinations.
        All four methods inpainted the same images from CelebA-HQ using the same masks.
        We used \patch in setup \cfg{c}.
    }
    \label{fig:cross-generator-localization}
\end{figure}

We evaluate to what extent localization methods trained on a particular generation method (\eg, diffusion, GAN) generalize to samples produced by a different one.
To this end we inpaint the CelebA-HQ dataset (using the same masks as before) with three other approaches:
Latent Diffusion Model (LDM) \cite{rombach2022high},
LaMa \cite{suvorov2022resolution} and
Pluralistic \cite{zheng2019pluralistic}.
For the Repaint--P2 dataset we use a subset of 9k samples to match the samples from the other approaches (see Table~\ref{tab:datasets}).
We train the \patch method in a fully supervised setting (scenario \cfg{c}) on each of the four datasets as well as combinations of those (using three out of the four datasets).
The evaluation is carried on all four inpainted test sets.

\begin{figure}%
    \centering
    \def\hf{38pt}
    \footnotesize
    \setlength{\tabcolsep}{1.5pt}
    \begin{tabular}{cccccc}
        & real
        & mask
        &
        &
        &
        \\
        & \includegraphics[align=c,height=\hf]{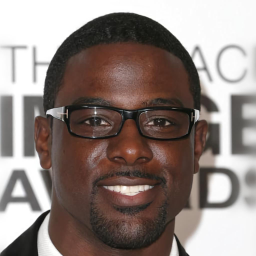}
        & \includegraphics[align=c,height=\hf]{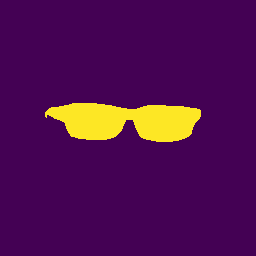}
        &
        & 
        & 
        \vspace{5pt} \\
        &
        & trained on
        & trained on
        & trained on
        & trained on
        \\
        & inpainted
        & repaint 
        & ldm 
        & lama    
        & pluralistic
        \\
        \rotatebox[origin=c]{90}{repaint}
        & \includegraphics[align=c,height=\hf]{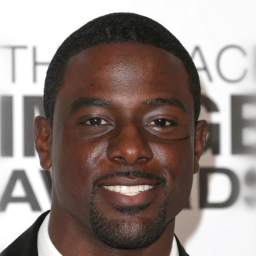} 
        & \includegraphics[align=c,height=\hf]{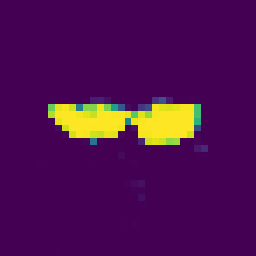} 
        & \includegraphics[align=c,height=\hf]{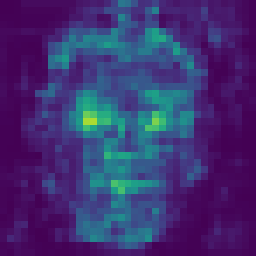} 
        & \includegraphics[align=c,height=\hf]{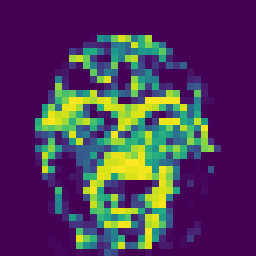} 
        & \includegraphics[align=c,height=\hf]{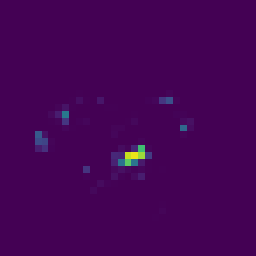}
        \vspace{3pt} \\
        \rotatebox[origin=c]{90}{ldm}
        & \includegraphics[align=c,height=\hf]{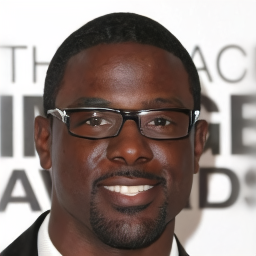} 
        & \includegraphics[align=c,height=\hf]{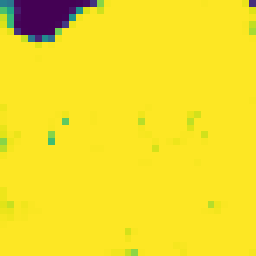} 
        & \includegraphics[align=c,height=\hf]{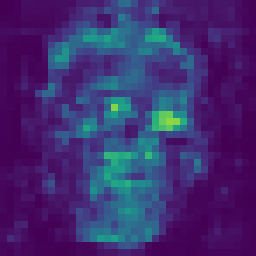} 
        & \includegraphics[align=c,height=\hf]{imgs/cross-generator/12551-2-pred-lama-01-repaint-p2-celebahq-9k-test.png} 
        & \includegraphics[align=c,height=\hf]{imgs/cross-generator/12551-2-pred-pluralistic-01-repaint-p2-celebahq-9k-test.png}
        \vspace{3pt} \\
        \rotatebox[origin=c]{90}{lama}
        & \includegraphics[align=c,height=\hf]{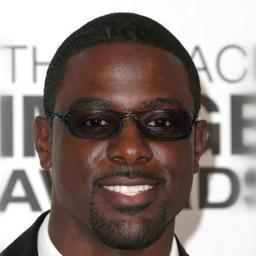}          
        & \includegraphics[align=c,height=\hf]{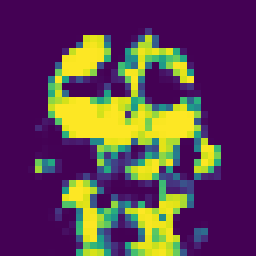}
        & \includegraphics[align=c,height=\hf]{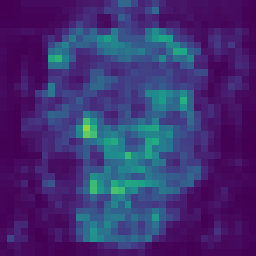}
        & \includegraphics[align=c,height=\hf]{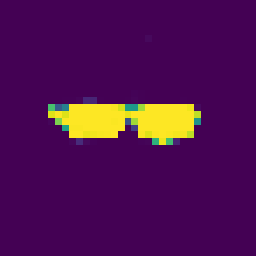}
        & \includegraphics[align=c,height=\hf]{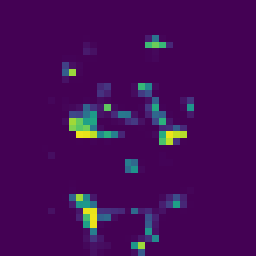}
        \vspace{3pt} \\
        \rotatebox[origin=c]{90}{pluralistic}
        & \includegraphics[align=c,height=\hf]{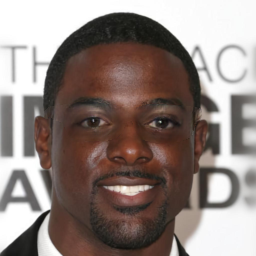}   
        & \includegraphics[align=c,height=\hf]{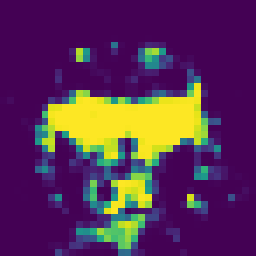}   
        & \includegraphics[align=c,height=\hf]{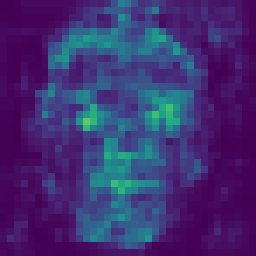}
        & \includegraphics[align=c,height=\hf]{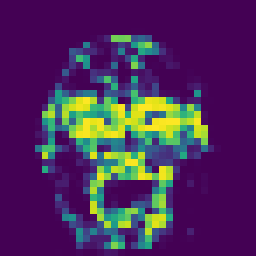}   
        & \includegraphics[align=c,height=\hf]{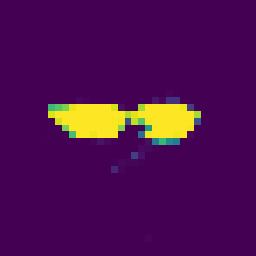}
        \\
    \end{tabular}
    \caption{%
        Qualitative results for the cross-generator evaluation using \patch trained in setup \cfg{c}.
        We observe the difficulty in generalization across generators (off-diagonal predictions) and the fact that local manipulations induced by LDM are challenging to identify (second image on the main diagonal).}
    
    \label{fig:cross-generator-qualitative}
\end{figure}

The results for the 32 train--test combinations are given in Figure \ref{fig:cross-generator-localization},
while qualitative results are shown in Figure \ref{fig:cross-generator-qualitative}.
We observe that localization works generally very well as long we test on data generated from the same model (main diagonal in the left plot).
However, LDM is an exception:
localization in LDM-manipulated images is more difficult since the inpainting is carried in the latent space and the decoding step ``hides'' the traces of the latent manipulation,
akin to how image processing steps degrade detection performance \cite{Wang_2020_CVPR}.

When we evaluate on data coming from a different generator,
the performance drops sharply (off-diagonal entries in the left plot). 
The transfer performance between LaMa and Pluralistic is still decent, presumably due to the particularities of the encoder--decoder approach.
The diffusion model of Repaint is different from the two and makes the cross-generator transfer more challenging.
Still, it appears that the transfer from diffusion to autoencoders and GANs (38.0\% and 41.4\%, respectively) is easier than the other way around (10.3\% and 0.2\%, respectively);
a similar conclusion has been observed for detection \cite{ricker2022towards}.

Training on combinations of multiple datasets yields generally good performance on all the datasets involved at training (off-diagonal entries in the right plot).
However, we do not observe a generalization benefit by using more types of generators at training (diagonal entries in the right plot \textit{vs} off-diagonal entries in the left plot).
For example, training on all generators but LDM yields an IoU of 19.7\%, which is only marginally above 19.5\%, what is achieved by training only on Repaint.
For the other three generators, the performance is even slightly worse on combinations than the single best generator.

\subsection{Performance on unseen datasets}
\label{sec:other-methods}

In this section, we consider generalization in its most challenging form, by varying both the source dataset and the generation algorithm.
Consequently, we evaluate on a different dataset, COCO Glide \cite{guillaro2022trufor}, which consists of 512 locally-edited images using a text-guided diffusion-based model.
Additionally, we present results of five other existing localization methods
\cite{wu2019mantra,cozzolino2019noiseprint,liu2022pscc,guillaro2022trufor,guo2023hierarchical}, which were pre-trained on different datasets,
and compare their performance on our own Repaint--P2/CelebA-HQ, as well as on COCO Glide.
For a comparison to \patch, we also fine-tune the PSCC method \cite{liu2022pscc} in setup \cfg{c} on the Repaint--P2/CelebA-HQ data.
The results are shown in Table~\ref{tab:comparison-pretrained}.

We observe that the generalization performance is modest on either of the two datasets:
the best out-of-domain performance on Repaint--P2/CelebA-HQ is 23.1\%, obtained by TruFor,
while on COCO Glide is 33.3\%, obtained by PSCC.
Even methods that have shown to generalize (TruFor \cite{guillaro2022trufor}) or that have been trained specifically on diffusion images (HiFi-Net \cite{guo2023hierarchical}) have difficulties on out-of-domain datasets.

\patch shows competitive results (second best in terms of IoU on COCO Glide),
even if it was trained solely on faces.
Interestingly, this is not the case for PSCC.
While PSCC obtains top performance in-domain, on Repaint--P2/CelebA-HQ, it struggles to generaralize to COCO Glide.
This behaviour suggests that overfitting is ocurring, which is not surprising given that the model capacity of PSCC (3.6M parameters) is an order of magnitude larger than the one of \patch (200k parameters).

\begin{table}
    \centering
    \small
    \begin{tabularx}{0.48\textwidth}{X rr rr}
        \toprule
        & \multicolumn{2}{c}{R-P2/CelebA}
        & \multicolumn{2}{c}{COCO Glide} \\
        \cmidrule(lr){2-3}
        \cmidrule(lr){4-5}
        Method & IoU & PBCA & IoU & PBCA  \\
        \midrule
        MantraNet \cite{wu2019mantra}             &      4.8 & \bf 81.9 &     25.1 &     79.8 \\
        Noiseprint \cite{cozzolino2019noiseprint} &     18.2 &     23.8 &     23.9 &     29.0 \\
        PSCC \cite{liu2022pscc}                   &     14.3 &     66.5 & \bf 33.3 &     80.6 \\
        TruFor \cite{guillaro2022trufor}          & \bf 23.1 &     81.3 &.    29.2 & \bf 81.4 \\
        HiFi-Net \cite{guo2023hierarchical}       &      0.0 &     81.0 &      2.6 &      3.2 \\
        \midrule
        \multicolumn{5}{l}{\textit{Methods trained on Repaint--P2/CelebA-HQ in setup \cfg{c}}} \\
        PSCC \cite{liu2022pscc}                   & \deemph{89.0} & \deemph{98.8} &  13.3 & 18.4 \\
        \patch                                    & \deemph{84.5} & \deemph{98.7} & 30.8 & 64.8 \\
        \bottomrule
    \end{tabularx}
    \caption{%
        Evaluation of pretrained localization models on our Repaint--P2/CelebA-HQ and the COCO Glide dataset \cite{guillaro2022trufor}.
        The grayed out results (\patch and PSCC on Repaint--P2/CelebA-HQ) are not directly comparable to those of other methods, since both \patch and PSCC are trained on Repaint--P2/CelebA-HQ.
        Qualitative results are available in the supplementary material.
    }
    \label{tab:comparison-pretrained}
\end{table}

\section{Conclusions}
\label{sec:conclusions}

In this paper, we investigate weakly-supervised localization in the context of diffusion-generated images of faces.
We propose a framework and a dataset that allows to systematically explore the importance of different factors in model performance, such as:
choice of detection method, level of supervision, dataset and type of generator used.
We design a series of experiments that progressively modify the training assumptions and showed that, to a certain extent, detection of local manipulations can be performed weakly supervised, even in the most restrictive scenarios.

We summarize our findings:
\begin{enumerate*}
    \item The patch-based method consistently outperforms the other two approaches (explanations or attention) across multiple settings and metrics.
    \item The detection performance in one of the weakly-supervised settings (image label \& partial manipulations) is strong across all detection methods, suggesting that partially-manipulated images can be used for training deepfake classifiers.
    \item Among the three types of factors (supervision, dataset, generator method), supervision seems to have a lesser impact (at least for the best performing method, \patch), while the generator impacts the most.
    \item Localization of manipulations for latent diffusion models is very challenging even in the most optimistic scenario.
\end{enumerate*}

We believe that these findings can fuel research into weakly-supervised localization of deepfake manipulations with possible extensions to general-content images and to other types of local manipulations,
such as face-swap, local enhancements or facial pose transfer obtained with DDPMs.

\mypar{Acknowledgements.}
This work was supported in part by European Union’s HORIZON-CL4-2021-HUMAN-01 research
and innovation program under grant agreement No. 101070190 ``AI4Trust''.

{\small
\bibliographystyle{ieee_fullname}
\bibliography{egbib}
}

\end{document}


\title{%
Weakly-supervised deepfake localization in diffusion-generated images\\
{\it (Supplementary material)}}

\maketitle

\section{Dataset details}
\label{sec:supp-dataset-details}

Table \ref{tab:sample_setups} presents the dataset information
for each of the three setups (\cfg{a}, \cfg{b}, \cfg{c})
in terms of the number of samples and
their provenance for each split (train, validation, test).
These details are relevant for the experiments in \S5.1 and \S5.2 in the main paper.

These data splits are built from real images (coming from either CelebA-HQ or FFHQ)
and fake images (generated by either P2 or Repaint--P2, which were trained on either CelebA-HQ or FFHQ).
For the weakly-supervised scenarios (\cfg{a} and \cfg{b}) we train on 9k real and 9k fake images,
the fake images being generated by P2.
For the fully-supervised scenario we use the same numbers of locally-inpainted samples for
both Repaint--P2/CelebaHQ and Repaint--P2/FFHQ: 30K train and 3K validation samples.

Our evaluation is always carried on data derived from CelebA-HQ
and even for the detection task we use partially-manipulated images (Repaint--P2/CelebA-HQ),
since our focus is weakly-supervised localization.

\section{Additional qualitative results}

\begin{figure*}
    \def\hf{42pt}
    \footnotesize
    \begin{center}        
    \setlength{\tabcolsep}{1.5pt}
        \begin{tabular}{cccccccccccccc}
 & real
            & inpainted
            & mask
            & &
            & real
            & inpainted
            & mask
            & &
            & real
            & inpainted
            & mask
            \\
            & \includegraphics[align=c,height=\hf]{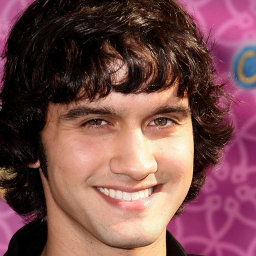}
            & \includegraphics[align=c,height=\hf]{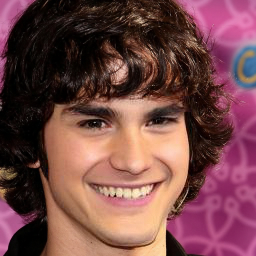}
            & \includegraphics[align=c,height=\hf]{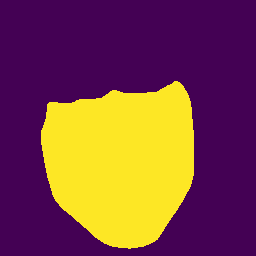}
            & &
            & \includegraphics[align=c,height=\hf]{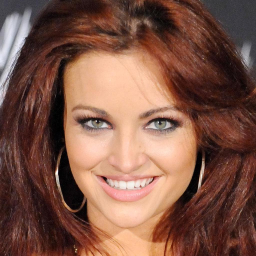}
            & \includegraphics[align=c,height=\hf]{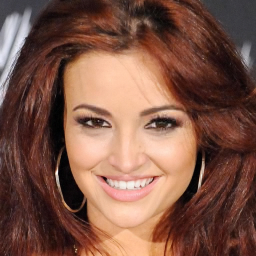}
            & \includegraphics[align=c,height=\hf]{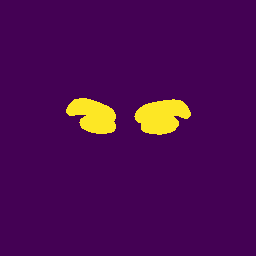}
            & &
            & \includegraphics[align=c,height=\hf]{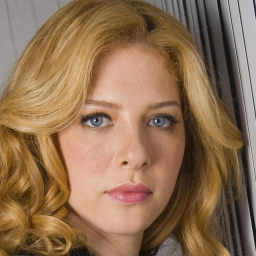}
            & \includegraphics[align=c,height=\hf]{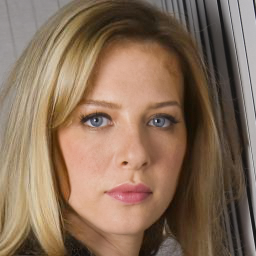}
            & \includegraphics[align=c,height=\hf]{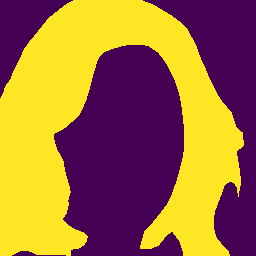}
            \vspace{3pt}
            \\
            & GC
            & PT
            & AT
            & &
            & GC
            & PT
            & AT
            & &
            & GC
            & PT
            & AT
            \\
            \cfg{a} 
            & \includegraphics[align=c,height=\hf]{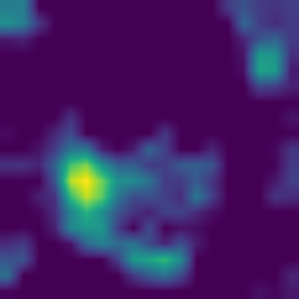}
            & \includegraphics[align=c,height=\hf]{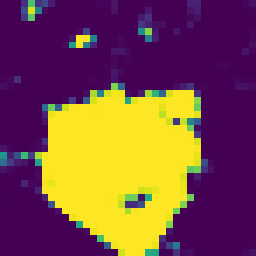}
            & \includegraphics[align=c,height=\hf]{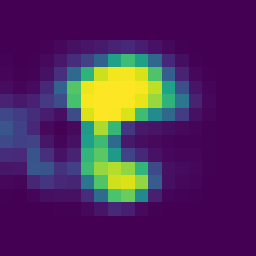}
            & &
            & \includegraphics[align=c,height=\hf]{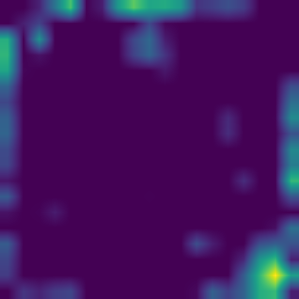}
            & \includegraphics[align=c,height=\hf]{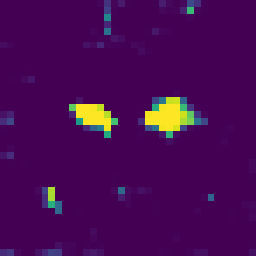}
            & \includegraphics[align=c,height=\hf]{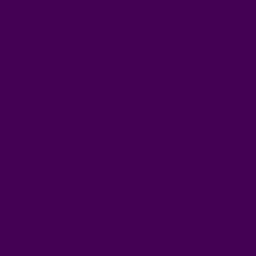}
            & &
            & \includegraphics[align=c,height=\hf]{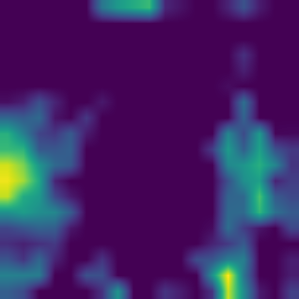}
            & \includegraphics[align=c,height=\hf]{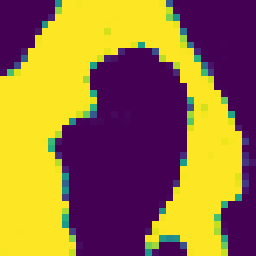}
            & \includegraphics[align=c,height=\hf]{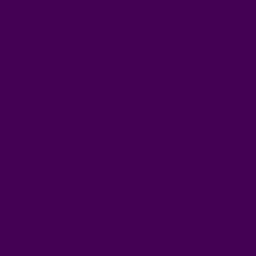}
            \vspace{3pt}
            \\
            \cfg{b} 
            & \includegraphics[align=c,height=\hf]{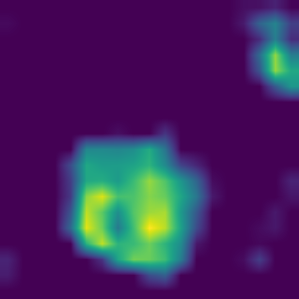}
            & \includegraphics[align=c,height=\hf]{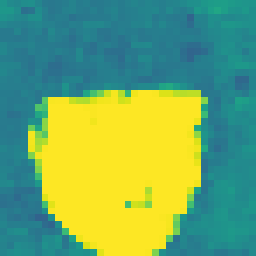}
            & \includegraphics[align=c,height=\hf]{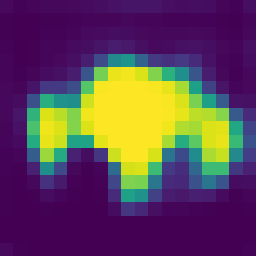}
            & &
            & \includegraphics[align=c,height=\hf]{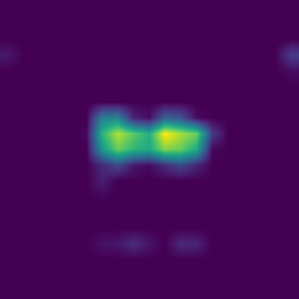}
            & \includegraphics[align=c,height=\hf]{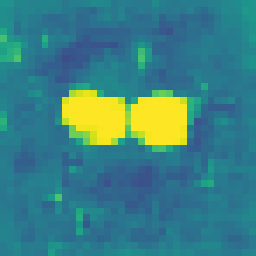}
            & \includegraphics[align=c,height=\hf]{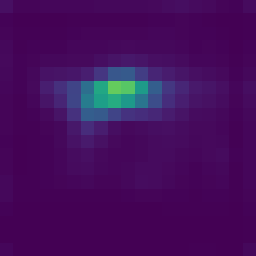}
            & &
            & \includegraphics[align=c,height=\hf]{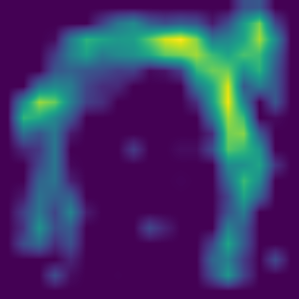}
            & \includegraphics[align=c,height=\hf]{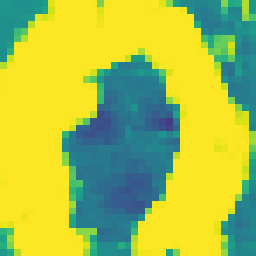}
            & \includegraphics[align=c,height=\hf]{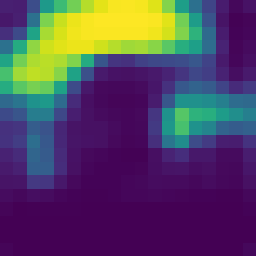}
            \vspace{3pt}
            \\  
            \cfg{c} 
            & \includegraphics[align=c,height=\hf]{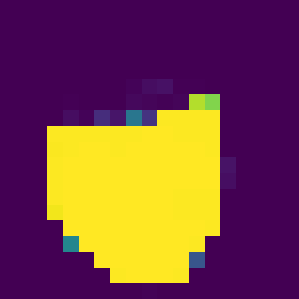}
            & \includegraphics[align=c,height=\hf]{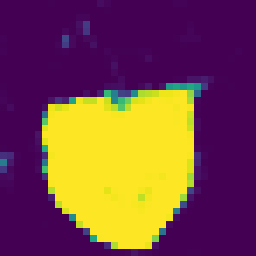}
            & \includegraphics[align=c,height=\hf]{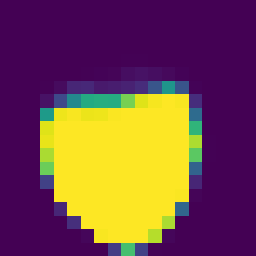}
            & &
            & \includegraphics[align=c,height=\hf]{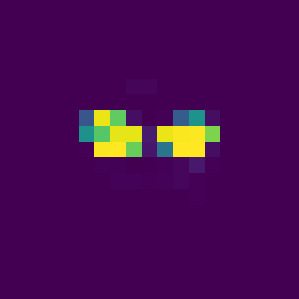}
            & \includegraphics[align=c,height=\hf]{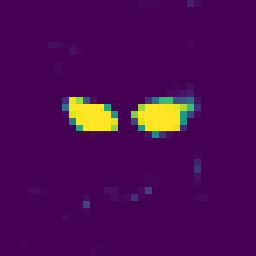}
            & \includegraphics[align=c,height=\hf]{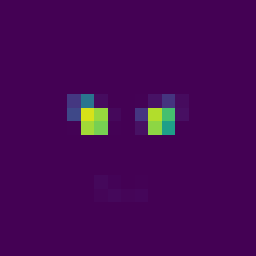}
            & &
            & \includegraphics[align=c,height=\hf]{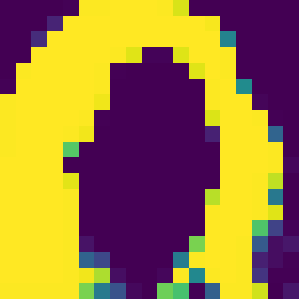}
            & \includegraphics[align=c,height=\hf]{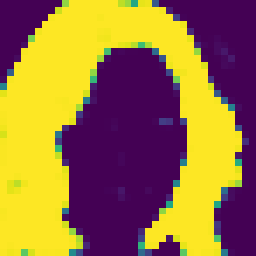}
            & \includegraphics[align=c,height=\hf]{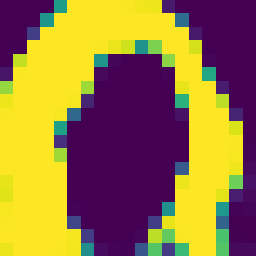}
            \vspace{3pt}
            \\
            \\
\hline
        \\
            & real
            & inpainted
            & mask
            & &
            & real
            & inpainted
            & mask
            & &
            & real
            & inpainted
            & mask
            \\
            & \includegraphics[align=c,height=\hf]{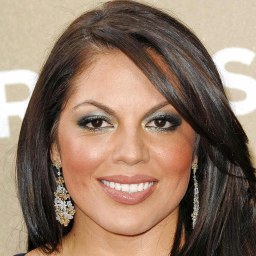}
            & \includegraphics[align=c,height=\hf]{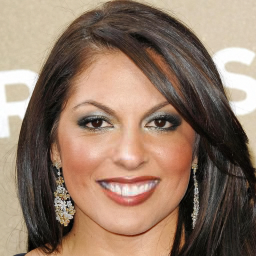}
            & \includegraphics[align=c,height=\hf]{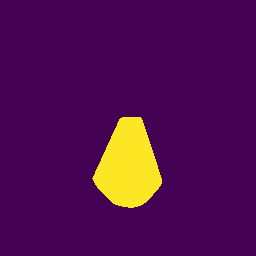}
            & &
            & \includegraphics[align=c,height=\hf]{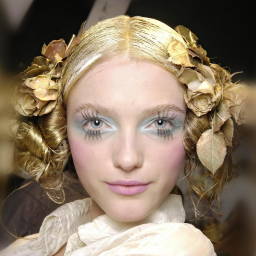}
            & \includegraphics[align=c,height=\hf]{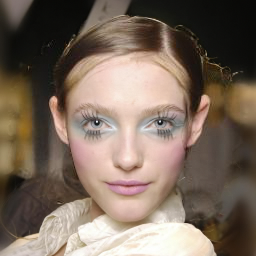}
            & \includegraphics[align=c,height=\hf]{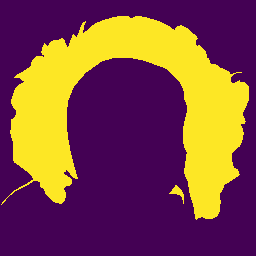}
            & &
            & \includegraphics[align=c,height=\hf]{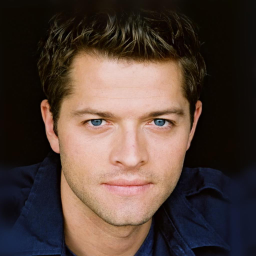}
            & \includegraphics[align=c,height=\hf]{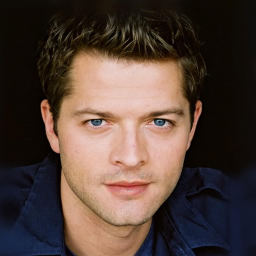}
            & \includegraphics[align=c,height=\hf]{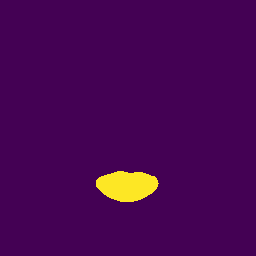}
            \vspace{3pt}
            \\
            & GC
            & PT
            & AT
            & &
            & GC
            & PT
            & AT
            & &
            & GC
            & PT
            & AT
            \\
            \cfg{a} 
            & \includegraphics[align=c,height=\hf]{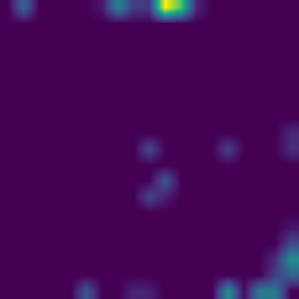}
            & \includegraphics[align=c,height=\hf]{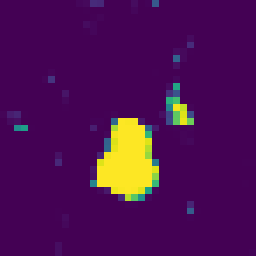}
            & \includegraphics[align=c,height=\hf]{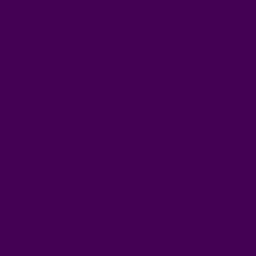}
            & &
            & \includegraphics[align=c,height=\hf]{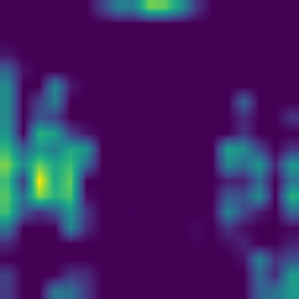}
            & \includegraphics[align=c,height=\hf]{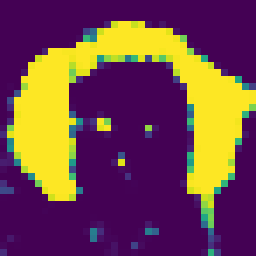}
            & \includegraphics[align=c,height=\hf]{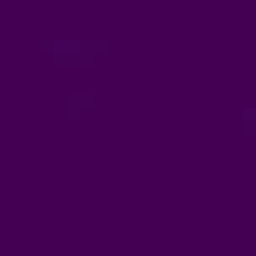}
            & &
            & \includegraphics[align=c,height=\hf]{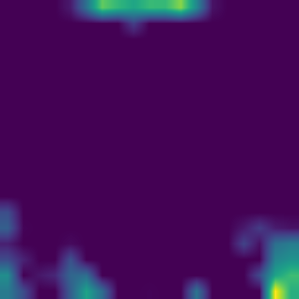}
            & \includegraphics[align=c,height=\hf]{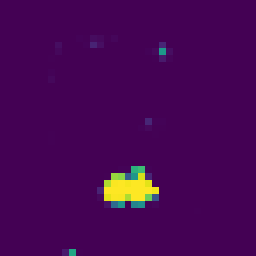}
            & \includegraphics[align=c,height=\hf]{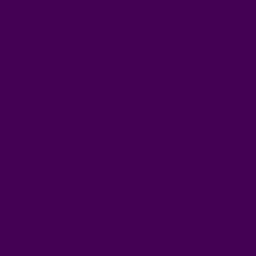}
            \vspace{3pt}
            \\
            \cfg{b} 
            & \includegraphics[align=c,height=\hf]{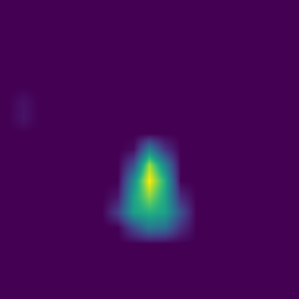}
            & \includegraphics[align=c,height=\hf]{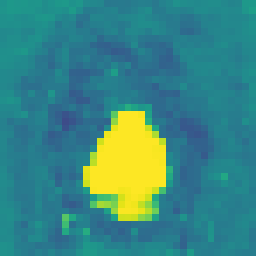}
            & \includegraphics[align=c,height=\hf]{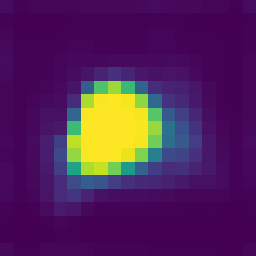}
            & &
            & \includegraphics[align=c,height=\hf]{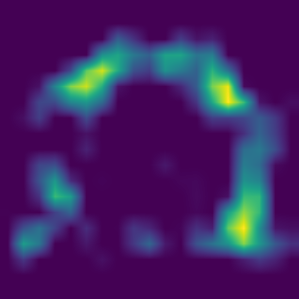}
            & \includegraphics[align=c,height=\hf]{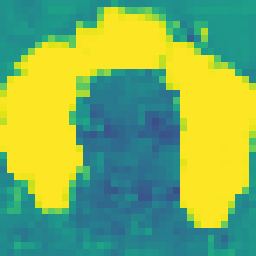}
            & \includegraphics[align=c,height=\hf]{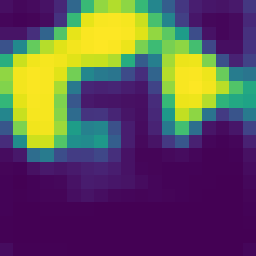}
            & &
            & \includegraphics[align=c,height=\hf]{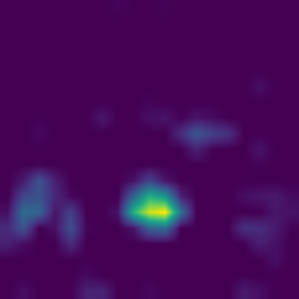}
            & \includegraphics[align=c,height=\hf]{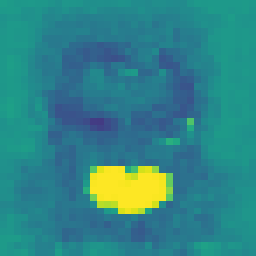}
            & \includegraphics[align=c,height=\hf]{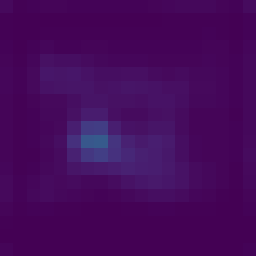}
            \vspace{3pt}
            \\  
            \cfg{c} 
            & \includegraphics[align=c,height=\hf]{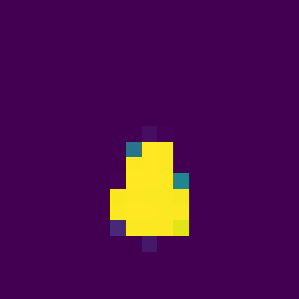}
            & \includegraphics[align=c,height=\hf]{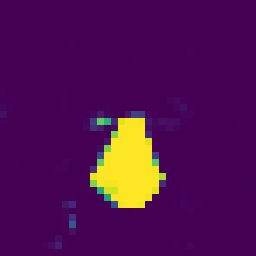}
            & \includegraphics[align=c,height=\hf]{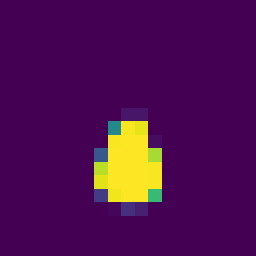}
            & &
            & \includegraphics[align=c,height=\hf]{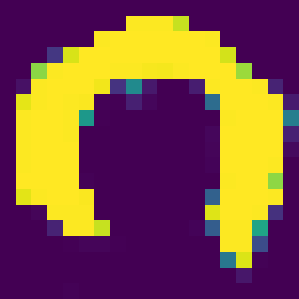}
            & \includegraphics[align=c,height=\hf]{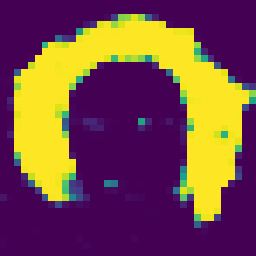}
            & \includegraphics[align=c,height=\hf]{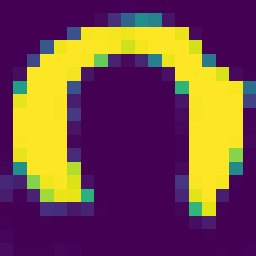}
            & &
            & \includegraphics[align=c,height=\hf]{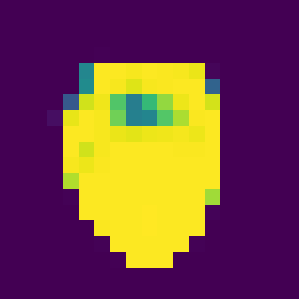}
            & \includegraphics[align=c,height=\hf]{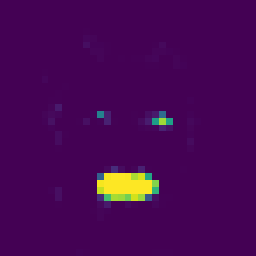}
            & \includegraphics[align=c,height=\hf]{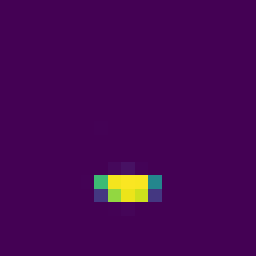}
            \vspace{3pt}\\
            \\

        \end{tabular}            
    \end{center}
    \caption[Localization visual results]{%
       Soft localization maps produced by the three proposed approaches using different level of supervision.
       \patch can accurately detect the manipulations after having seen only fully generated fake images (scenario \cfg{a}) or locally-inpainted images with only image-level supervision (scenario \cfg{b}).
       Both \att and \grad struggle in scenarios \cfg{a} and \cfg{b}.
       All methods recover the manipulated region in the fully supervised scenario, \cfg{c}.
       This suggests that operating at a patch level is better suited for recovering local manipulations than either using a \grad or \att.
    }
    \label{fig:maps-sup1}
\end{figure*}

\begin{figure*}
    \def\hf{34pt}
    \footnotesize
    \begin{center}        
    \setlength{\tabcolsep}{1.5pt}
        \begin{tabular}{c|cc|cc|cc|cc|cc|cc|}
            & inpainted
            & mask
            & inpainted
            & mask 
            & inpainted
            & mask 
            & inpainted
            & mask
            & inpainted
            & mask 
            & inpainted
            & mask 
            \\
            & \includegraphics[align=c,height=\hf]{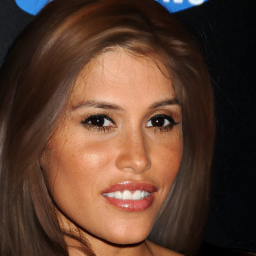}
            & \includegraphics[align=c,height=\hf]{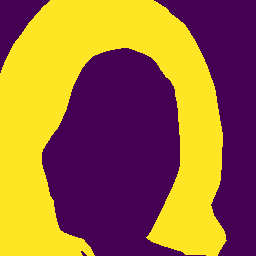}
            & \includegraphics[align=c,height=\hf]{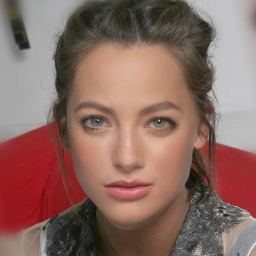}
            & \includegraphics[align=c,height=\hf]{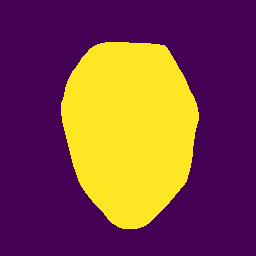}
            & \includegraphics[align=c,height=\hf]{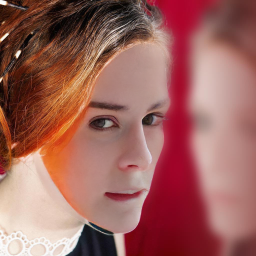}
            & \includegraphics[align=c,height=\hf]{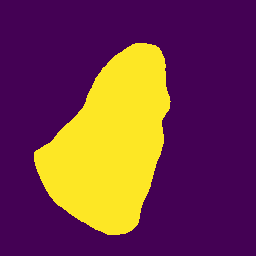}

            & \includegraphics[align=c,height=\hf]{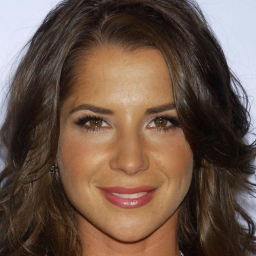}
            & \includegraphics[align=c,height=\hf]{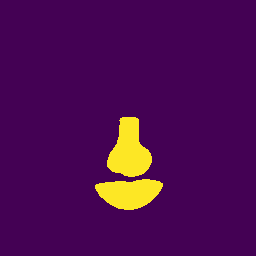}
            & \includegraphics[align=c,height=\hf]{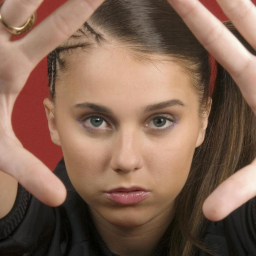}
            & \includegraphics[align=c,height=\hf]{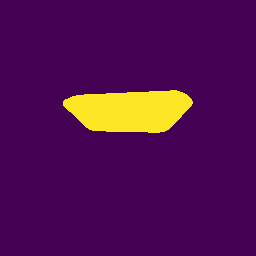}
            & \includegraphics[align=c,height=\hf]{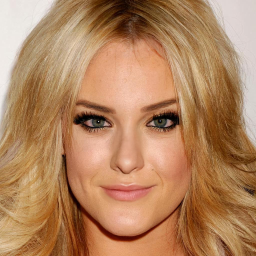}
            & \includegraphics[align=c,height=\hf]{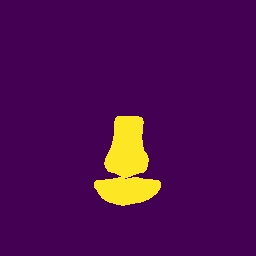}
            \vspace{3pt}
            \\
            & CelebA-HQ
            & FFHQ
            & CelebA-HQ
            & FFHQ
            & CelebA-HQ
            & FFHQ
            & CelebA-HQ
            & FFHQ
            & CelebA-HQ
            & FFHQ
            & CelebA-HQ
            & FFHQ
            \\
            \cfg{a} 
            & \includegraphics[align=c,height=\hf]{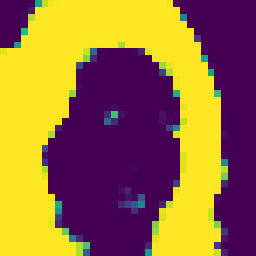}
            & \includegraphics[align=c,height=\hf]{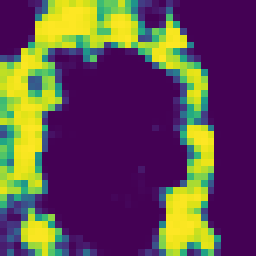}
            & \includegraphics[align=c,height=\hf]{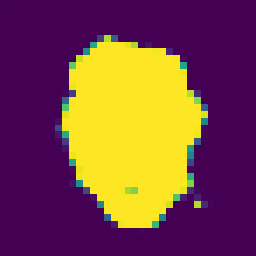}
            & \includegraphics[align=c,height=\hf]{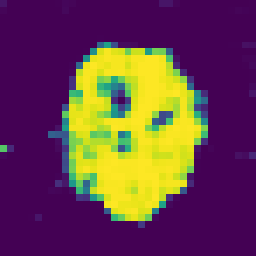}
            & \includegraphics[align=c,height=\hf]{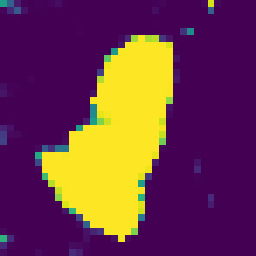}
            & \includegraphics[align=c,height=\hf]{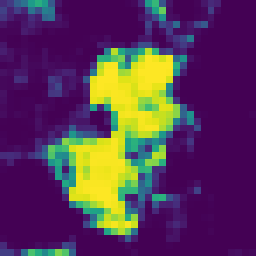}
 \vspace{3pt}
            & \includegraphics[align=c,height=\hf]{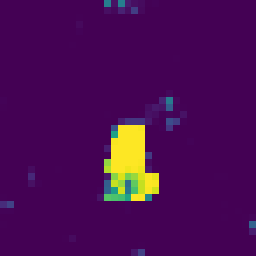}
            & \includegraphics[align=c,height=\hf]{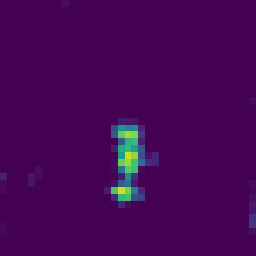}
            & \includegraphics[align=c,height=\hf]{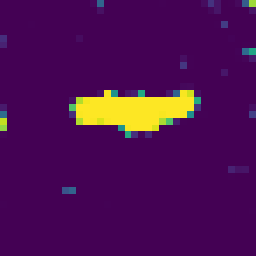}
            & \includegraphics[align=c,height=\hf]{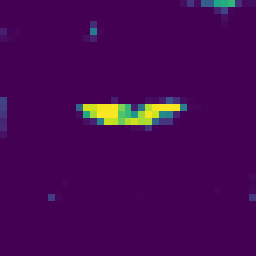}
            & \includegraphics[align=c,height=\hf]{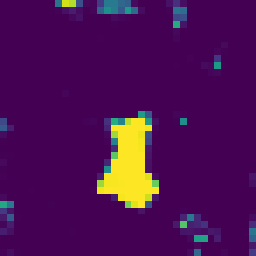}
            & \includegraphics[align=c,height=\hf]{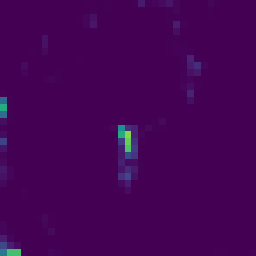}
            \vspace{3pt}
            \\
            \cfg{b} 
            & \includegraphics[align=c,height=\hf]{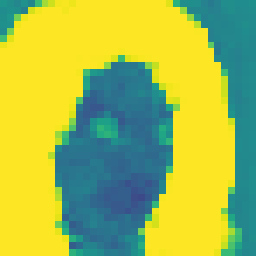}
            & \includegraphics[align=c,height=\hf]{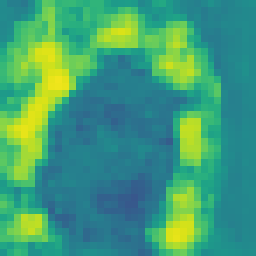}
            & \includegraphics[align=c,height=\hf]{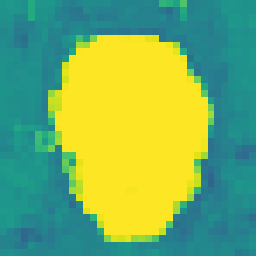}
            & \includegraphics[align=c,height=\hf]{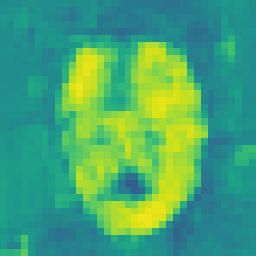}
            & \includegraphics[align=c,height=\hf]{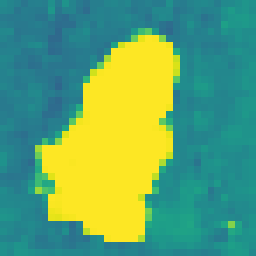}
            & \includegraphics[align=c,height=\hf]{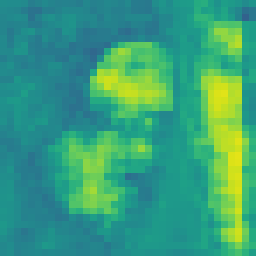}
 \vspace{3pt}
            & \includegraphics[align=c,height=\hf]{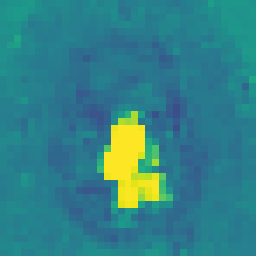}
            & \includegraphics[align=c,height=\hf]{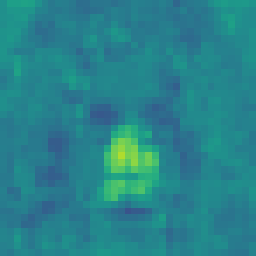}
            & \includegraphics[align=c,height=\hf]{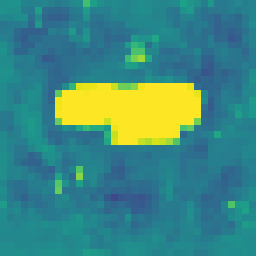}
            & \includegraphics[align=c,height=\hf]{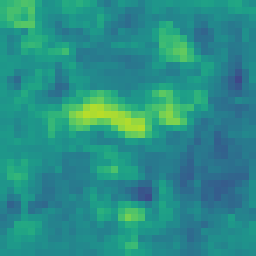}
            & \includegraphics[align=c,height=\hf]{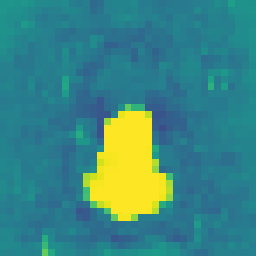}
            & \includegraphics[align=c,height=\hf]{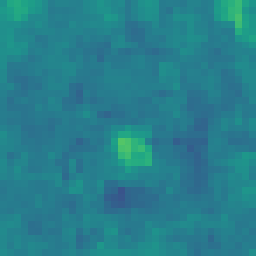}
            \vspace{3pt}
            \\  
            \cfg{c} 
            & \includegraphics[align=c,height=\hf]{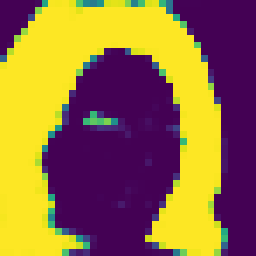}
            & \includegraphics[align=c,height=\hf]{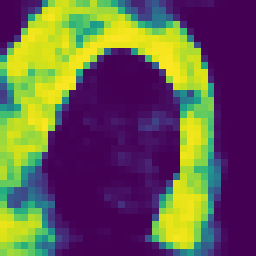}
            & \includegraphics[align=c,height=\hf]{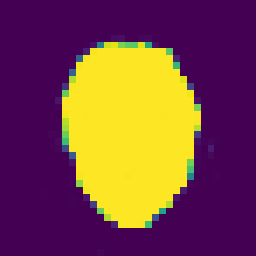}
            & \includegraphics[align=c,height=\hf]{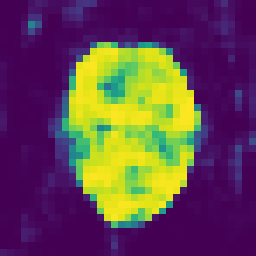}
            & \includegraphics[align=c,height=\hf]{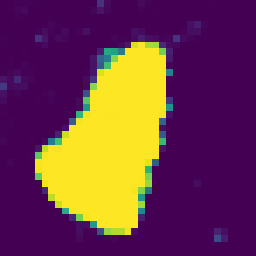}
            & \includegraphics[align=c,height=\hf]{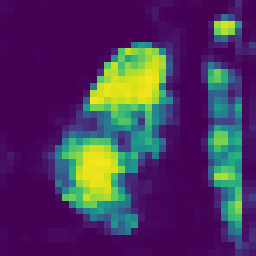}
 \vspace{3pt}
            & \includegraphics[align=c,height=\hf]{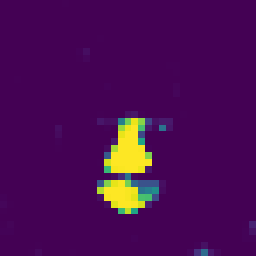}
            & \includegraphics[align=c,height=\hf]{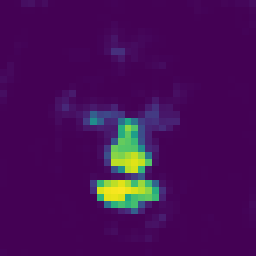}
            & \includegraphics[align=c,height=\hf]{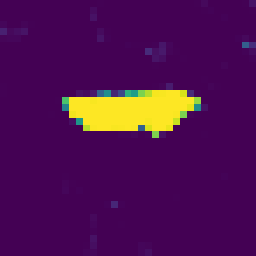}
            & \includegraphics[align=c,height=\hf]{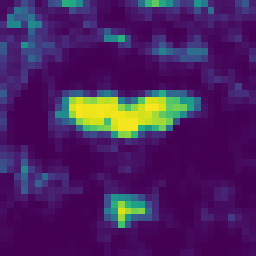}
            & \includegraphics[align=c,height=\hf]{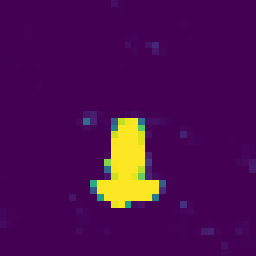}
            & \includegraphics[align=c,height=\hf]{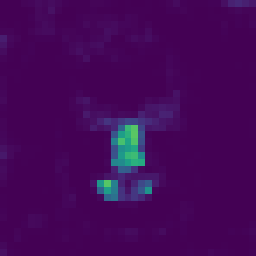}
            \vspace{3pt}
            \\
        \end{tabular}            
    \end{center}
    \caption[localization visual results]{%
        Soft localization maps when using the same and different source datasets for training and testing. For training we use data derived either form CelebA-HQ or FFHQ while for testing we use data derived from CelebA-HQ.
        With different training and testing source datasets the produced maps become less sharp and eroded, especially in the harder weakly supervised scenarios, \cfg{a} and \cfg{b}.
        Due to the noisy nature of the training in scenario \cfg{b} the separation between real and fake regions is dimmed. 
    }
    \label{fig:maps_sup2}
\end{figure*}

We present additional visual results that paint a more complete image of the performance of the proposed models in different training setups. Firstly, in Figure \ref{fig:maps-sup1} we show visual results for all three methods \patch, \att, \grad, on the three identified training scenarios: \cfg{a}, \cfg{b}, \cfg{c}.
We notice that \patch performs the best in all setups. 
In Figure \ref{fig:maps_sup2} we show additional results when using the same and different datasets for training and for testing. The level of performance degradation is larger for smaller masks. 

\section{Comparison to other pretrained models}

\begin{table}[hb!]
\centering
\small
\begin{tabular}{c cc rr rr}
\toprule
& & &
\multicolumn{2}{c}{IoU (\%)} &
\multicolumn{2}{c}{PBCA (\%)} \\
\cmidrule(lr){4-5}
\cmidrule(lr){6-7}
& sup. & generator & 
\multicolumn{1}{c}{SC} & \multicolumn{1}{c}{FT} & 
\multicolumn{1}{c}{SC} & \multicolumn{1}{c}{FT} \\
\midrule                           
\deemph{\cfg{a}} & label & full    &     10.7 &  6.0 & 71.5 & 79.8 \\ %
\deemph{\cfg{b}} & label & partial &       -- & 18.4 &   -- & 21.3 \\ %
\deemph{\cfg{c}} & label & partial &     89.0 & 93.9 & 98.8 & 99.5 \\ %
\bottomrule
\end{tabular}
\caption{%
Localization performance on Repaint--P2/CelebA-HQ by initializing the training of PSCC either from scratch (SC) or by finetuning the pretrained model (FT).
}
\label{tab:finetune}
\end{table}

\begin{figure*}[htb!]
    \def\hf{40pt}
    \scriptsize 
    \begin{center}        
        \setlength{\tabcolsep}{2pt}
        \begin{tabular}{cc|cccccc}
            inpainted & mask & MantraNet & Noiseprint & PSCC & TruFor & HiFi-Net & \patch \\
            \includegraphics[align=c,height=\hf]{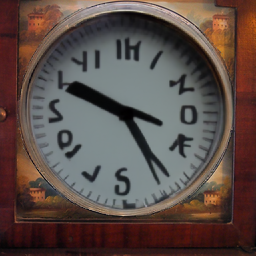}
            & \includegraphics[align=c,height=\hf]{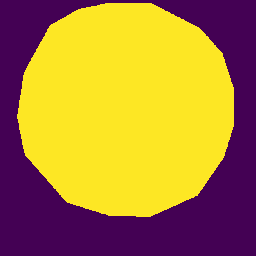}
            & \includegraphics[align=c,height=\hf]{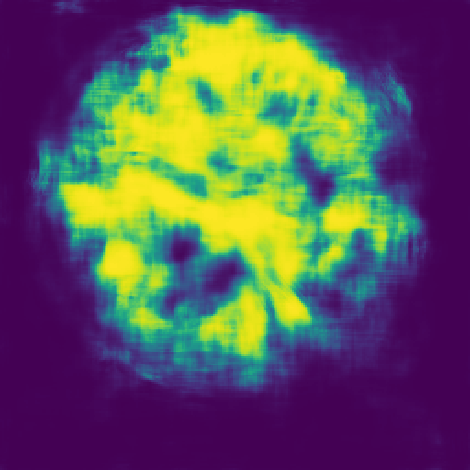}
            & \includegraphics[align=c,height=\hf]{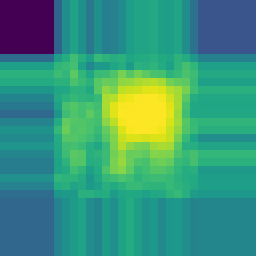}
            & \includegraphics[align=c,height=\hf]{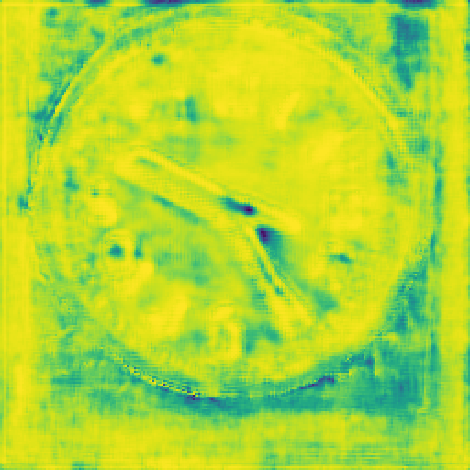}
            & \includegraphics[align=c,height=\hf]{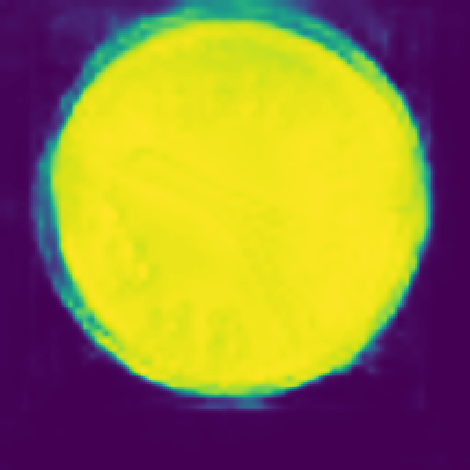}
            & \includegraphics[align=c,height=\hf]{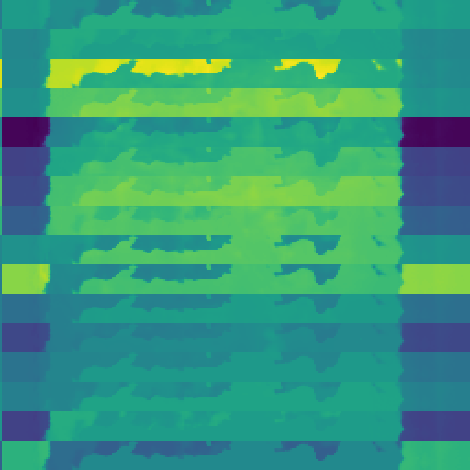}
            & \includegraphics[align=c,height=\hf]{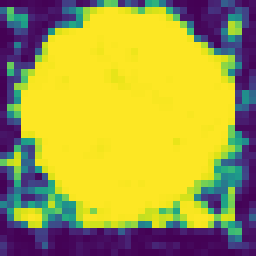}
            \\
            \includegraphics[align=c,height=\hf]{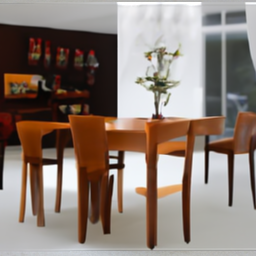}
            & \includegraphics[align=c,height=\hf]{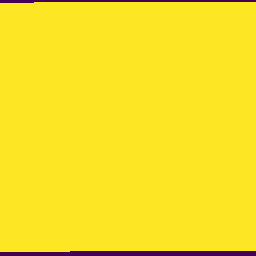}
            & \includegraphics[align=c,height=\hf]{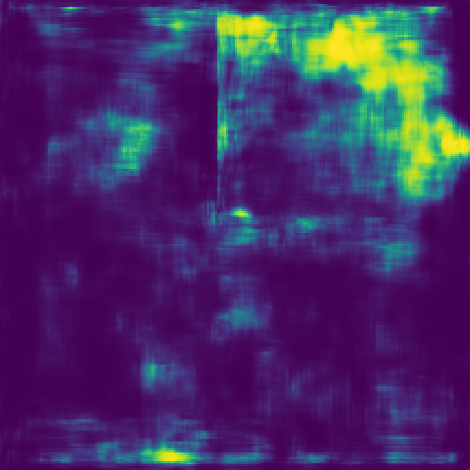}
            & \includegraphics[align=c,height=\hf]{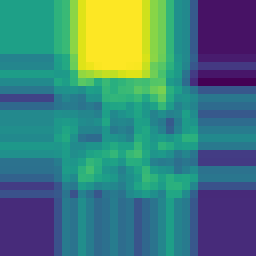}
            & \includegraphics[align=c,height=\hf]{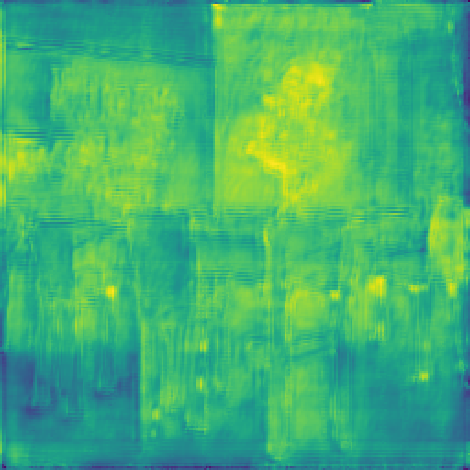}
            & \includegraphics[align=c,height=\hf]{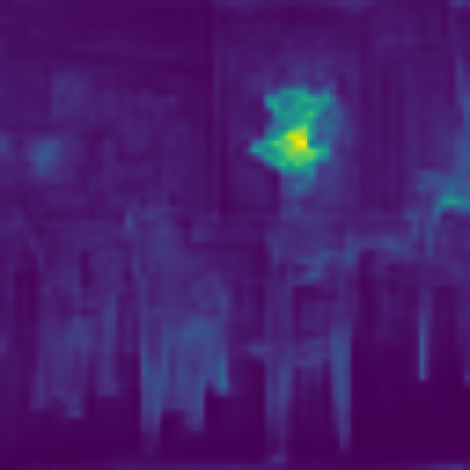}
            & \includegraphics[align=c,height=\hf]{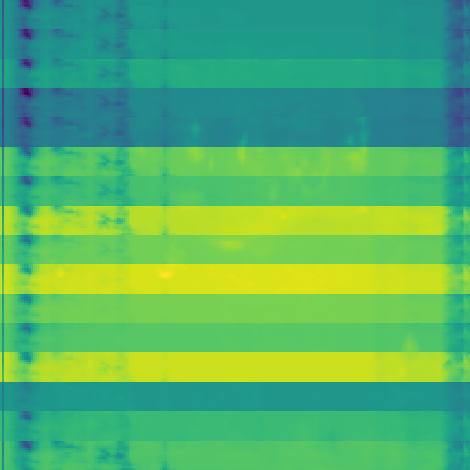}
            & \includegraphics[align=c,height=\hf]{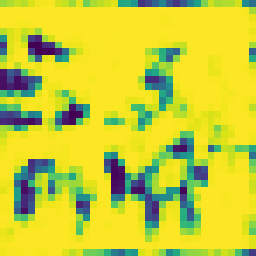}
            \\
            \includegraphics[align=c,height=\hf]{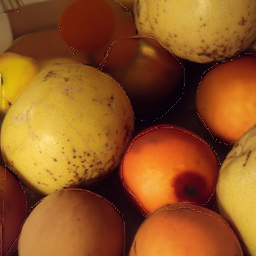}
            &\includegraphics[align=c,height=\hf]{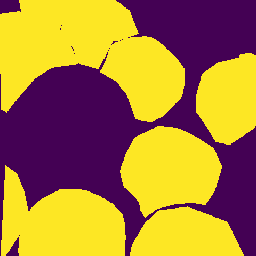}
            &\includegraphics[align=c,height=\hf]{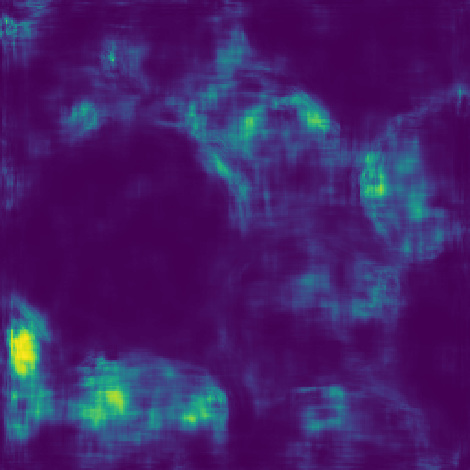}
            &\includegraphics[align=c,height=\hf]{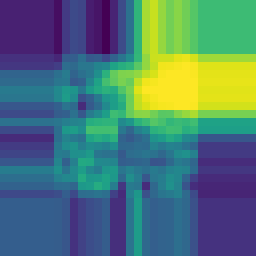}
            &\includegraphics[align=c,height=\hf]{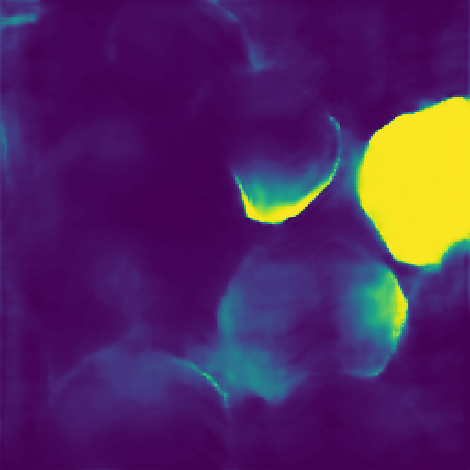}
            &\includegraphics[align=c,height=\hf]{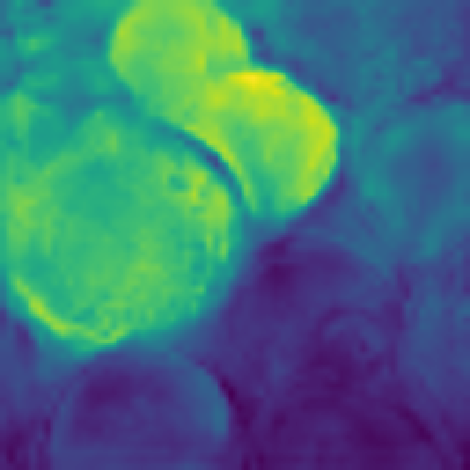}
            &\includegraphics[align=c,height=\hf]{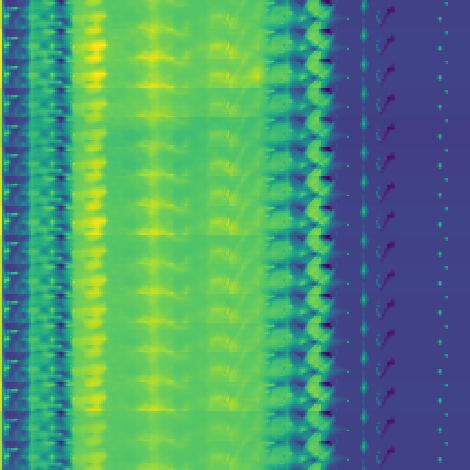}
            &\includegraphics[align=c,height=\hf]{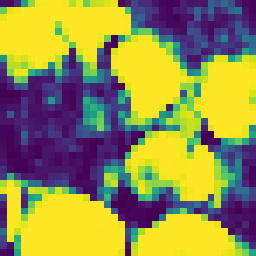}
            \\
            \includegraphics[align=c,height=\hf]{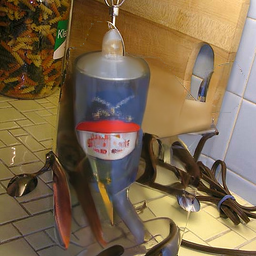}
            &\includegraphics[align=c,height=\hf]{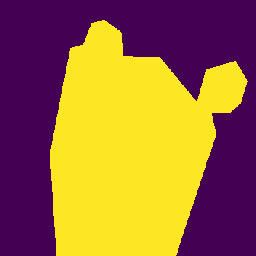}
            &\includegraphics[align=c,height=\hf]{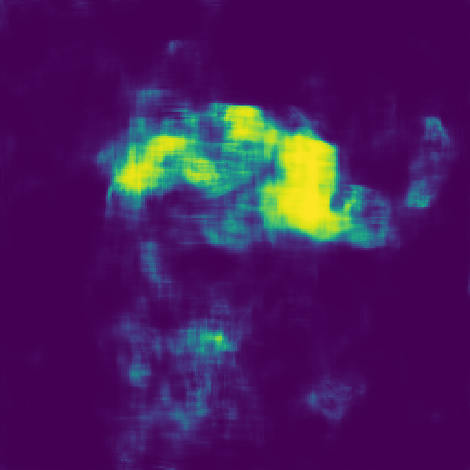}
            &\includegraphics[align=c,height=\hf]{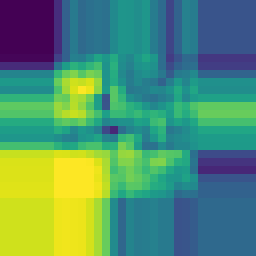}
            &\includegraphics[align=c,height=\hf]{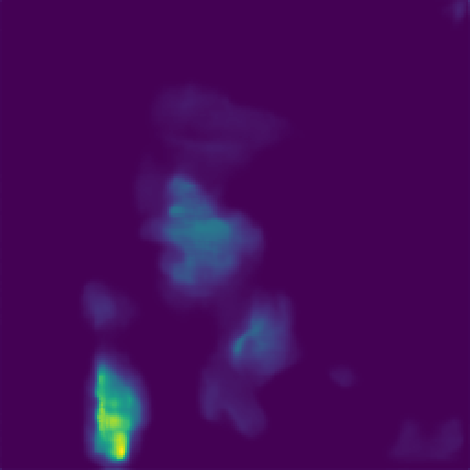}
            &\includegraphics[align=c,height=\hf]{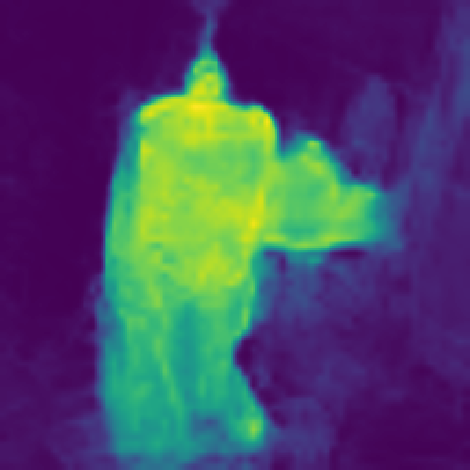}
            &\includegraphics[align=c,height=\hf]{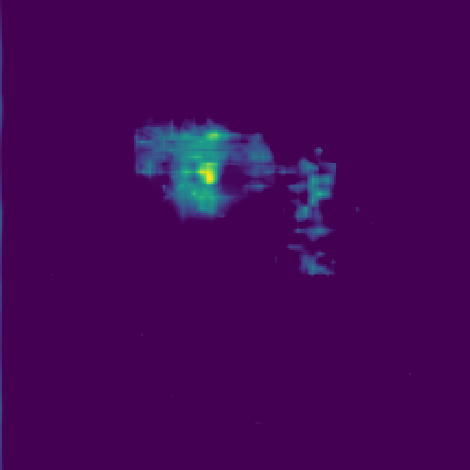}
            &\includegraphics[align=c,height=\hf]{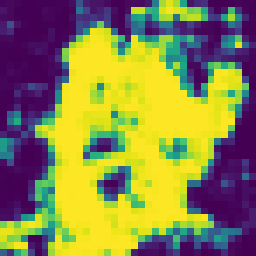}
            \\
            \includegraphics[align=c,height=\hf]{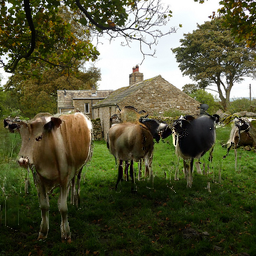}
            &\includegraphics[align=c,height=\hf]{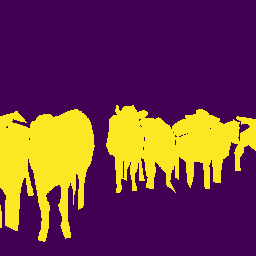}
            &\includegraphics[align=c,height=\hf]{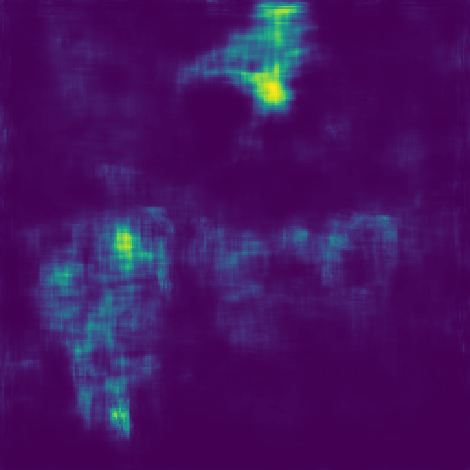}
            &\includegraphics[align=c,height=\hf]{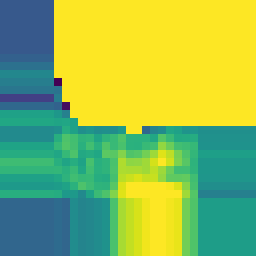}
            &\includegraphics[align=c,height=\hf]{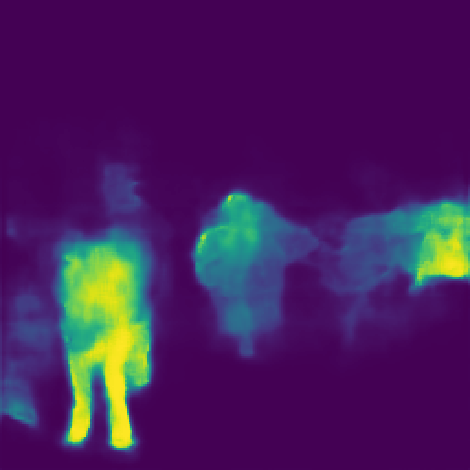}
            &\includegraphics[align=c,height=\hf]{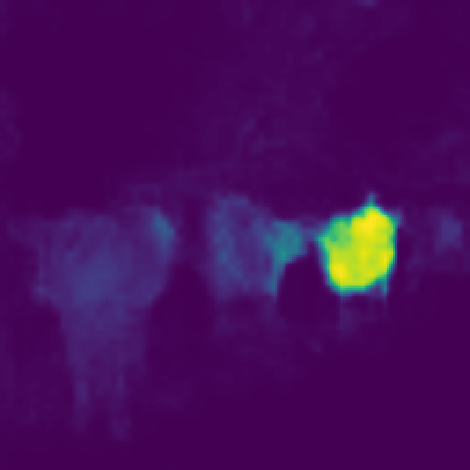}
            &\includegraphics[align=c,height=\hf]{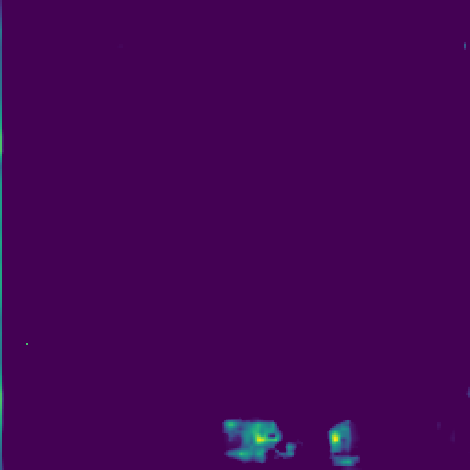}
            &\includegraphics[align=c,height=\hf]{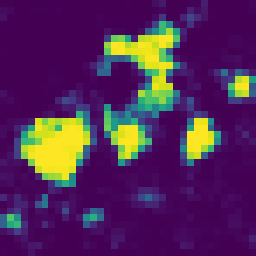}
            \\
            \includegraphics[align=c,height=\hf]{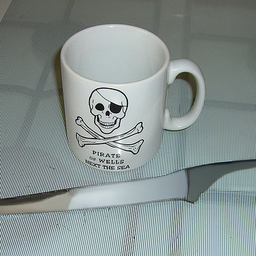}
            &\includegraphics[align=c,height=\hf]{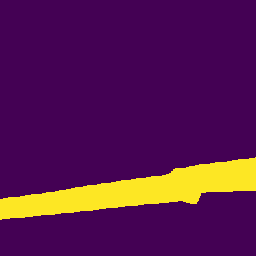}
            &\includegraphics[align=c,height=\hf]{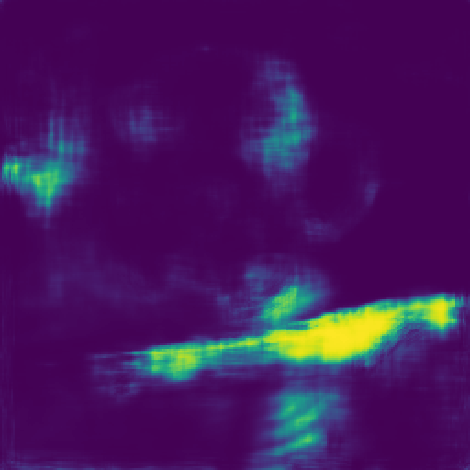}
            &\includegraphics[align=c,height=\hf]{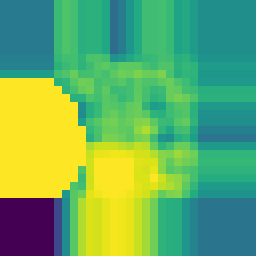}
            &\includegraphics[align=c,height=\hf]{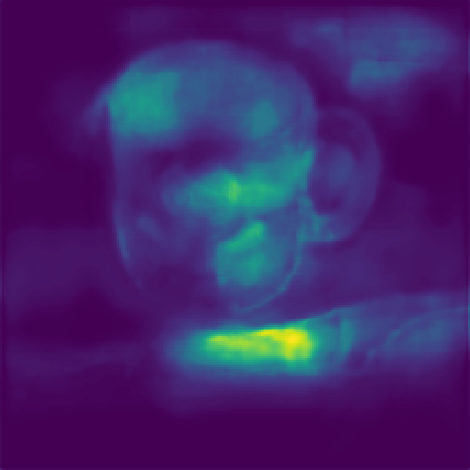}
            &\includegraphics[align=c,height=\hf]{imgs/supplementary/coco_glide/mantra/glide_inpainting_val2017_2592_up.png}
            &\includegraphics[align=c,height=\hf]{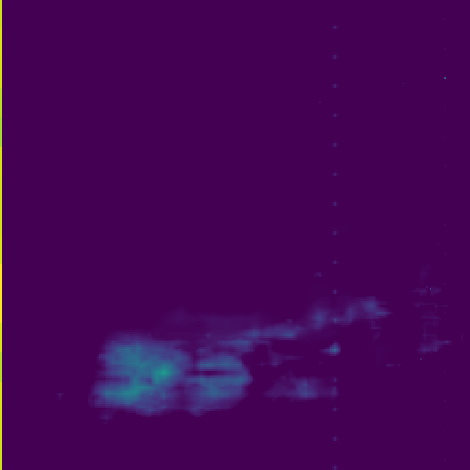}
            &\includegraphics[align=c,height=\hf]{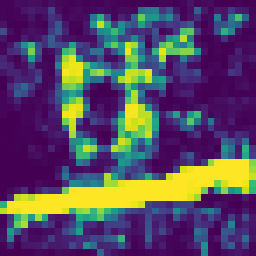}
            \\
            \includegraphics[align=c,height=\hf]{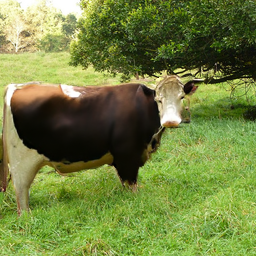}
            &\includegraphics[align=c,height=\hf]{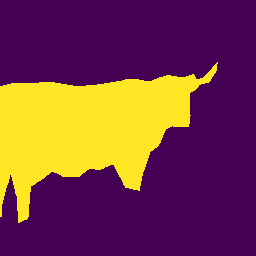}
            &\includegraphics[align=c,height=\hf]{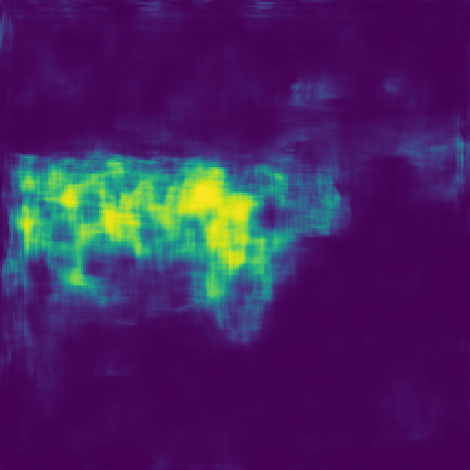}
            &\includegraphics[align=c,height=\hf]{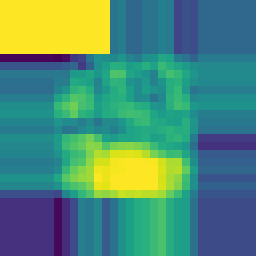}
            &\includegraphics[align=c,height=\hf]{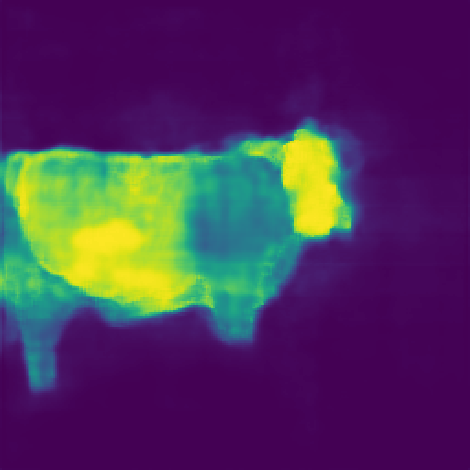}
            &\includegraphics[align=c,height=\hf]{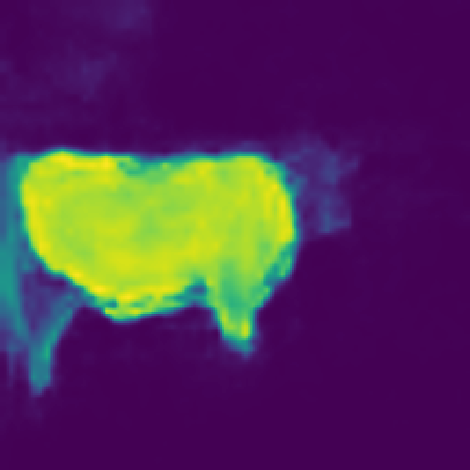}
            &\includegraphics[align=c,height=\hf]{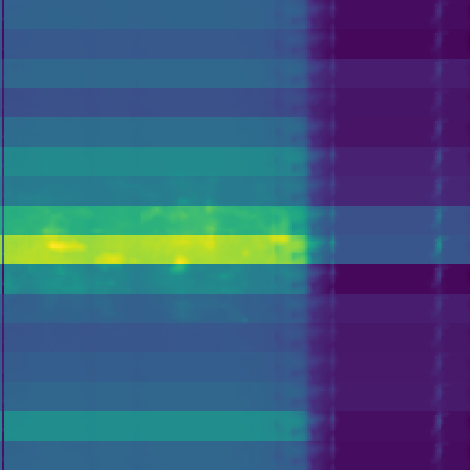}
            &\includegraphics[align=c,height=\hf]{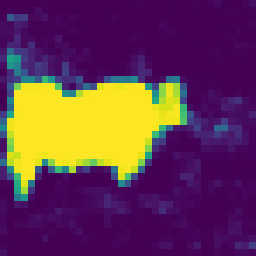}
        \end{tabular}
    \end{center}
    \caption{%
        Visual results obtained with five pre-trained methods: MantraNet, Noiseprint, PSCC, TruFor, HiFi-Net and \patch on COCO Glide dataset. 
        For these visualizations all methods are trained fully-supervised.%
    }
    \label{fig:coco_glide}
\end{figure*}

We compare to five pretrained models for detection and fully-supervised localization (see Figure \ref{fig:coco_glide}).
MantraNet and PSCC are trained on data forged with copy-move, splicing, removal and enhancement operations.
Noiseprint relies on noise-removal techniques and learns to distinguish whether the input patches come from the same source (have similar noise residual patterns).
HiFi-Net and TruFor are recent approaches (CVPR'23). 
The former is trained on diffusion and GAN images (with local and full manipulations) to produce hierarchical attributes.
The latter is an improved version of Noiseprint, which is also trained on images from more recent manipulation techniques (GAN). 

Visual results in Figure \ref{fig:coco_glide} indicate that Noiseprint and Hi-Fi net struggle the most to recover the inpainted regions. The activations obtained with MantraNet seem reasonable, but the network lacks the confidence and hence the small numerical results under a standard threshold of 0.5. PSCC and TruFor generally seem to find the manipulated region but they tend to under or over-segment it. Similarly, \patch is mostly correct in localizing the fake area but lacks precision. Unlike other methods, \patch has only been trained to localize forgeries of faces.  The competitive results obtained with \patch  on COCO Glide dataset suggest that it is a suitable method to perform analysis in more challenging weakly-supervised scenarios.

\section{Additional results with PSCC}

Table~\ref{tab:finetune} presents results for PSCC trained in all three scenarios.
To ensure a fair comparison, we have trained the PSCC method similarly to \patch.
In particular,
(i) we have initialized the model from scratch (random weights), and
(ii) for scenarios \cfg{a} and \cfg{b}, which provide only a label,
we have broadcasted the label to a image-sized matrix to obtain the mask,
which is needed as target.
However, in the inherent noisy training setup of configuration \cfg{b}, we have observed that the model did not converge.
Instead, we were able to train in this scenario by starting from the checkpoint provided by the authors.
In the paper, we report results by training from scratch.

\begin{table*}
\tabcolsep 4pt
\centering
\small
\newcommand{\na}{{\color{gray!30}N/A}}
\begin{tabular}{c cc | lr | lr | lr | lr | lr | lr | lr}
    \toprule
    & & & \multicolumn{4}{c|}{train} & \multicolumn{4}{c|}{valid}  & \multicolumn{2}{c|}{test loc.} & \multicolumn{4}{c}{test det.} \\
    & & &
    \multicolumn{2}{c}{real} & \multicolumn{2}{c|}{fake} &
    \multicolumn{2}{c}{real} & \multicolumn{2}{c|}{fake} &
    \multicolumn{2}{c|}{fake} & 
    \multicolumn{2}{c}{real} & \multicolumn{2}{c}{fake} \\
    & sup. & generator             
    & \multicolumn{1}{c}{src.} & \multicolumn{1}{c|}{num.} & \multicolumn{1}{c}{src.} & \multicolumn{1}{c|}{num.}
    & \multicolumn{1}{c}{src.} & \multicolumn{1}{c|}{num.} & \multicolumn{1}{c}{src.} & \multicolumn{1}{c|}{num.}
    & \multicolumn{1}{c}{src.} & \multicolumn{1}{c|}{num.} 
    & \multicolumn{1}{c}{src.} & \multicolumn{1}{c|}{num.} & \multicolumn{1}{c}{src.} & \multicolumn{1}{c}{num.} \\
    \midrule
    \deemph{\cfg{a}} & label & full    & $d$  & 9k  & P2/$d$   &  9k & $d$ & 900 & P2/$d$   & 900 & R.P2/CA & 8.5k & CA & 900 & R.P2/CA & 900 \\
    \deemph{\cfg{b}} & label & partial & $d$  & 9k  & R.P2/$d$ &  9k & $d$ & 900 & R.P2/$d$ & 900 & R.P2/CA & 8.5k & CA & 900 & R.P2/CA & 900 \\
    \deemph{\cfg{c}} & mask  & partial &  \na & \na & R.P2/$d$ & 30k & \na & \na & R.P2/$d$ & 3k  & R.P2/CA & 8.5k & \na & \na & \na & \na \\
    \bottomrule
\end{tabular}
\caption{%
     Datasets used for each of our setups in terms of number of samples (num.) and their provenance (src.) for each of the real and fake parts as well as for each of the splits.
     We use $d$ to denote one of the two datasets (CelebA-HQ or FFHQ),
     while R.P2 stands for Repaint--P2 and CA for CelebA-HQ.
     Note that the evaluation is always carried out on data derived from CelebA-HQ.
}
\label{tab:sample_setups}
\end{table*}

